\documentclass[10pt,journal,compsoc]{IEEEtran}
\usepackage{amsmath,amsfonts}
\usepackage[noend]{algorithmic}
\usepackage{algorithm}
\usepackage{array}
\usepackage[caption=false,font=normalsize,labelfont=sf,textfont=sf]{subfig}
\usepackage{textcomp}
\usepackage{stfloats}
\usepackage{url}
\usepackage{verbatim}
\usepackage{graphicx}
\usepackage[nocompress]{cite}
\usepackage{booktabs}
\usepackage{makecell}
\usepackage{float}
\usepackage{multirow}
\usepackage{placeins}
\usepackage{color, colortbl}
\usepackage{xcolor}

\definecolor{graytext}{gray}{0.5} % 定义灰色
\renewcommand{\algorithmiccomment}[1]{\hfill\parbox[t]{0.55\linewidth}{\raggedright\textcolor{graytext}{\# #1}}}

\definecolor{dgreen}{RGB}{1,150,74}
\usepackage[pagebackref=true,breaklinks=true,letterpaper=true,colorlinks,bookmarks=false,citecolor=citecolor]{hyperref}
\definecolor{citecolor}{HTML}{0071BC}

\newcommand\red[1]{\textcolor{red}{#1}}

\newcommand\up[1]{\textcolor{dgreen}{$\uparrow{#1}$}}

\def\name{FSF }
\def\namenospace{FSF}

\begin{document}

\title{Fully Sparse Fusion for 3D Object Detection}


\author{
Yingyan Li, 
Lue Fan, 
Yang Liu, 
Zehao Huang, 
Yuntao Chen, 
Naiyan Wang
and Zhaoxiang Zhang
\IEEEcompsocitemizethanks{
\IEEEcompsocthanksitem{
This work was supported in part by the National Key R\&D Program of China (No. 2022ZD0160102), the National Natural Science Foundation of China (No. U21B2042, No. 62072457), and in part by the 2035 Innovation Program of CAS.
}
\IEEEcompsocthanksitem{
Yingyan~Li, Lue~Fan, Yang~Liu and Zhaoxiang~Zhang are with Center for Research on Intelligent Perception and Computing (CRIPAC), 
State Key Laboratory of Multimodal Artificial Intelligence Systems (MAIS), Institute of Automation, Chinese Academy of Sciences (CASIA), Beijing 100190, China. 
E-mail: \{liyingyan2021, fanlue2019, liuyang2022, zhaoxiang.zhang\}@ia.ac.cn.}
\IEEEcompsocthanksitem{
Yingyan~Li, Lue~Fan and Zhaoxiang~Zhang are with the School of Future Technology, University of Chinese Academy
of Sciences (UCAS), Beijing 100049, China. 
E-mail: \{liyingyan2021, fanlue2019, zhaoxiang.zhang\}@ia.ac.cn.}
\IEEEcompsocthanksitem{
Yuntao Chen and Zhaoxiang Zhang are with Centre for Artificial Intelligence and
Robotics, Hong Kong Institute of Science and Innovation, Chinese
Academy of Sciences (HKISI CAS), Hong Kong, China.
E-mail of Yuntao Chen: chenyuntao08@gmail.com
}
\IEEEcompsocthanksitem{
Zehao~Huang and Naiyan~Wang are with TuSimple, Beijing 100020,
China. E-mail: \{zehaohuang18, winsty\}@gmail.com.}
}% <-this % stops an unwanted space
}
% \author{Lue Fan,
%         Yuxue Yang,
%         Feng Wang,
%         Naiyan Wang,
%         and Zhaoxiang Zhang
% \IEEEcompsocitemizethanks{
% \IEEEcompsocthanksitem{Lue~Fan and Yuxue~Yang and Zhaoxiang~Zhang are with Center for Research on Intelligent Perception and Computing (CRIPAC), National Laboratory of Pattern Recognition (NLPR), Institute of Automation, Chinese Academy of Sciences (CASIA), Beijing 100190, China. E-mail: \{fanlue2019, yangyuxue2023, zhaoxiang.zhang\}@ia.ac.cn.}%\protect\\
% % note need leading \protect in front of \\ to get a newline within \thanks as
% % \\ is fragile and will error, could use \hfil\break instead.
% \IEEEcompsocthanksitem{
%           Feng~Wang and Naiyan~Wang are with TuSimple, Beijing 100020,
%           China. E-mail: \{feng.wff, winsty\}@gmail.com.}
% }% <-this % stops an unwanted space
% % \thanks{Manuscript received April 19, 2005; revised August 26, 2015.}
% }

\markboth{Journal of \LaTeX\ Class Files,~Vol.~14, No.~8, August~2021}%
{Shell \MakeLowercase{\textit{et al.}}: A Sample Article Using IEEEtran.cls for IEEE Journals}


\IEEEtitleabstractindextext{%
\begin{abstract}
Currently prevalent multi-modal 3D detection methods rely on dense detectors that usually use dense Bird's-Eye-View (BEV) feature maps. 
However, the cost of such BEV feature maps is quadratic to the detection range, making it not scalable for long-range detection. 
Recently, LiDAR-only fully sparse architecture has been gaining attention for its high efficiency in long-range perception.
In this paper, we study how to develop a multi-modal fully sparse detector.
Specifically, our proposed detector integrates the well-studied 2D instance segmentation into the LiDAR side, which is parallel to the 3D instance segmentation part in the LiDAR-only baseline. 
The proposed instance-based fusion framework maintains full sparsity while overcoming the constraints associated with the LiDAR-only fully sparse detector.
Our framework showcases state-of-the-art performance on the widely used nuScenes dataset, Waymo Open Dataset, and the long-range Argoverse 2 dataset. 
Notably, the inference speed of our proposed method under the long-range perception setting is 2.7$\times$ faster than that of other state-of-the-art multimodal 3D detection methods.
Code is released at \url{https://github.com/BraveGroup/FullySparseFusion}.
\end{abstract}

% % Note that keywords are not normally used for peerreview papers.
}

\maketitle

% \IEEEpeerreviewmaketitle

\section{Introduction}

\begin{figure}[!ht]
\vspace{6mm}
\centering
\includegraphics[width=0.85\columnwidth]{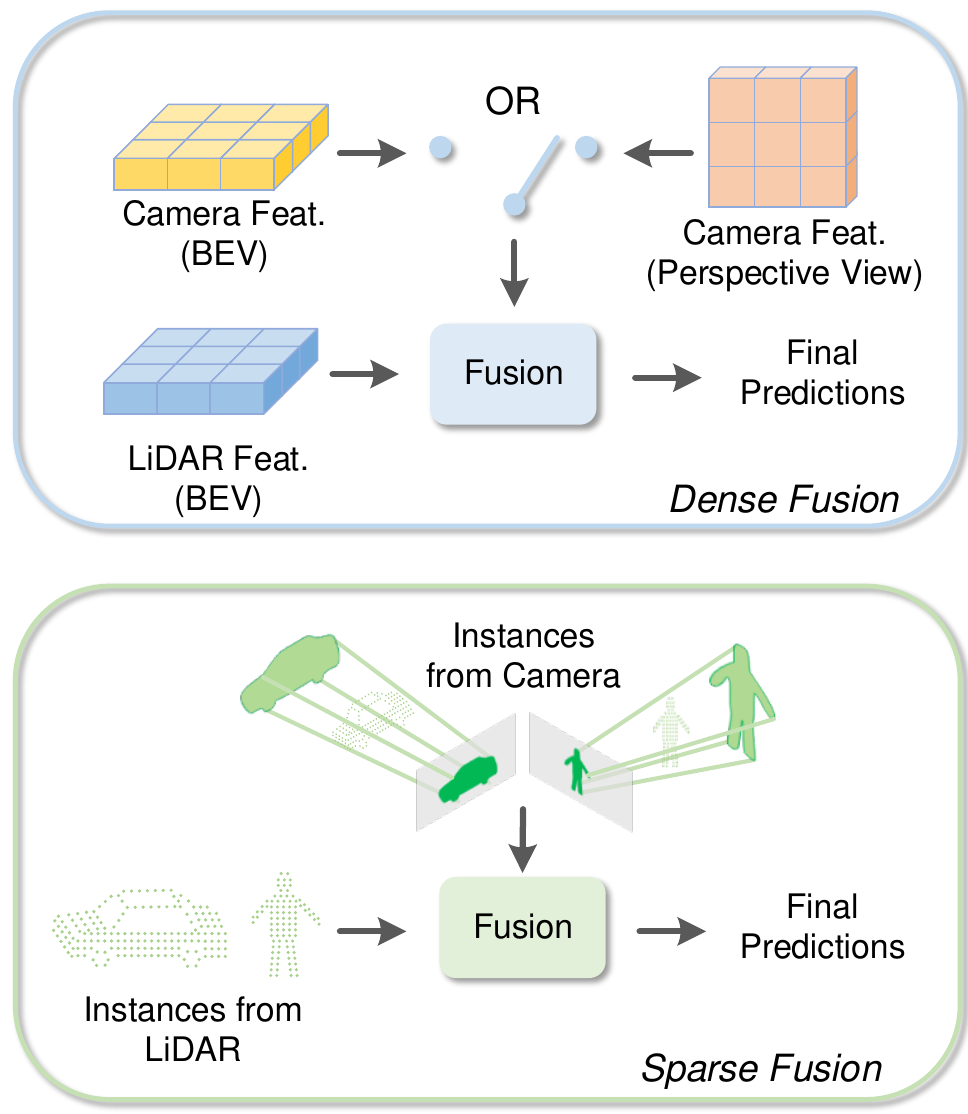}
\caption{
Comparison between dense fusion and sparse fusion. 
Dense fusion methods rely on dense BEV feature maps.
In contrast, our sparse fusion framework fuses two modalities at the instance level, requiring no dense feature maps.
}\label{fig:fig1}
\end{figure}

3D object detection serves as a crucial component in autonomous driving systems. 
Currently, LiDARs and cameras are the two main sensors used for detection. 
Combining these two sensors~\cite {bevfusion-damo, bevfusion-mit, transfusion} is popular in current perception system, which leverages precise depth of LiDARs and rich semantics of cameras.
Currently, the predominant multi-modal methods~\cite{bevfusion-damo, bevfusion-mit, transfusion} rely on dense detectors.
Dense detectors imply the detectors construct dense Bird's-Eye-View (BEV) feature maps for prediction as Fig.~\ref{fig:fig1} shows.
The size of these BEV feature maps increases quadratically with perception range, which leads to unaffordable costs in long-range detection. 

To solve this problem, we aim to design a multi-modal detector without introducing any dense BEV features.
Recently, several LiDAR-only fully sparse detectors~\cite{fsd, voxelnext, flatformer, fsd++} have emerged, where fully sparse implies that the whole architecture does not include any BEV feature map.
As a representative, Fully Sparse Detector (FSD)~\cite{fsd} demonstrates impressive efficiency and efficacy.
However, given FSD, how to build a fully sparse multimodal 3D detector that guarantees efficiency and effectiveness remains an unknown issue.
On the one hand, we need to ensure that the entire framework remains fully sparse, without introducing BEV features. 
On the other hand, we aim to effectively address the shortcomings of FSD.
FSD relies on clustering-based 3D instance segmentation, which makes it easy to ignore the foreground objects with few points, and hard to differentiate closely adjacent objects, leading to low object recall.

Therefore, we propose Fully Sparse Fusion (FSF), a detector concentrating on addressing the above two challenges from the perspective of instance-based fusion. 
First, we use the well-studied 2D instance segmentation model to yield 2D masks, which possess enhanced object recall due to the rich semantics and high resolution of images. 
Next, we transform each 2D instance mask into a 3D frustum, according to the intrinsic and extrinsic parameters of cameras.
The points contained within this frustum constitute a 3D instance corresponding to the mask.
These 3D instances from image modality are complementary to those 3D instances generated from LiDAR-based instance segmentation in FSD.
Finally, a bi-modal instance-based prediction module is leveraged to fuse these two kinds of instances and make final predictions.
This module is featured with a couple of sub-modules including instance feature extraction/interaction, instance shape alignment, and a prediction head.
Furthermore, due to the different shapes and spatial distributions of the bi-modal instances, conventional label assignment is insufficient. To this end, we propose a two-stage assignment to assign labels to the instances from two modalities.

The proposed framework offers multiple benefits.
Firstly, it is totally based on instances without introducing any dense BEV features, guaranteeing full sparsity.
Additionally, it gets the best of two worlds by seamlessly integrating well-studied 2D instance segmentation and 3D instance segmentation, mitigating the low recall of the LiDAR side and facilitating highly accurate object localization.
Due to incorporating instance-based information, FSF is superior to the semantic-only fusion approaches such as PointPainting~\cite{pointpainting} and PointAugmenting~\cite{pointaugmenting}.

We summarize our contributions as follows:
\begin{itemize}
    \item We propose Fully Sparse Fusion (FSF), a novel fully sparse multi-modality perception framework, which leverages instance-based fusion without any dense BEV feature maps. 
    \item FSF seamlessly integrates 2D instance segmentation and 3D instance segmentation by bi-modal instance generation and bi-modal instance-based prediction, making the best of two worlds.
    \item FSF achieves state-of-the-art performance in the nuScenes dataset~\cite{nus}, Waymo Open Dataset~\cite{wod}, and Argoverse 2 dataset~\cite{argo2}. Particularly, for the long-range detection in the Argoverse 2 dataset, FSF is 2.7 $\times$ faster than the previous state-of-the-art multi-modal detector.
\end{itemize}
\label{sec:intro}

\section{Related Work}
\subsection{Camera-based 3D detection}
In the early years of camera-based 3D detection, the focus is mainly on monocular 3D detection~\cite{m3d-rpn,autoshape,monoflex,gupnet,dcd, monodistill, badet}.
However, monocular 3D detection ignores the relationships between cameras.
To this end, more and more researchers are engaged in multi-view 3D detection.
Multi-view 3D detection~\cite{stereo-rcnn, liga-stereo, frustumformer} contains two main types of approaches: BEV-based and object-query-based. 
BEV-based methods~\cite{caddn, bevformer} lift 2D features to 3D~\cite{lss,bevdet4d,bevdepth} or using transformer-like structures~\cite{bevformer,bevformerv2,frustumformer} to construct BEV feature map for prediction.
On the contrary, object-query-based methods~\cite{detr3d,petr,mv2d} predict 3D bounding boxes from the features of object queries.

\subsection{LiDAR-based 3D detection}
Although camera-based 3D detection methods gain great improvements in the past few years, their performance is still inferior to LiDAR-based 3D detection methods since LiDAR points have accurate depths.
LiDAR-based methods~\cite{pointpillar, voxelnet} commonly transform irregular LiDAR point clouds into regular spaces, like 2D bird's-eye view or 3D voxels, to facilitate feature extraction.
PointPillars~\cite{pointpillar}, VoxelNet~\cite{voxelnet} and 3DFCN~\cite{3dfcn} employ dense 2D/3D convolution to extract features.
However, utilizing dense convolution introduces heavy computational costs.
To address this issue, SECOND~\cite{second} applies sparse convolution to 3D detection and inspires an increasing number of methods~\cite{centerpoint, pvrcnn, largekernel, focalspconv} adopting sparse convolution for feature extraction.
However, even with sparse convolution, most methods~\cite{centerpoint, cylinder3d} still require a dense feature map to alleviate the so-called ``center feature missing" issue~\cite{fsd}. 
Researchers have been dedicated to developing fully sparse methods to better utilize the sparsity nature of points. 
The pioneering works in this domain are PointNet-like methods~\cite{pointnet, pointnet++, 3dssd, votenet, pointrcnn}.
Also, range view~\cite{rangedet, fcos-lidar} requires no BEV feature map, though their performance is inferior. 
Recently, FSD~\cite{fsd} has been proposed and demonstrated faster speed and higher accuracy than previous methods. 
Our work is based on FSD.

\subsection{Multi-modal 3D detection}
% \fan{Divide this subsection into two subsections: dense fusion (BEVFusion) and sparse fusion (painting, mvp)}
The multi-modal 3D detection approaches can be classified into two types, namely dense and sparse, depending on whether it relies on the dense BEV feature map. 

\noindent{\textbf{Dense Fusion.}} As the representation of dense frameworks, BEVFusion~\cite{bevfusion-mit, bevfusion-damo} fuses the BEV feature maps generated from camera and LiDAR modalities. 
Another line of dense fusion frameworks~\cite{transfusion, 3d-cvf, avod, mv3d, uvtr, futr3d} relies on the BEV feature map to generate proposals. 
There is a concurrent work named SparseFusion~\cite{sparsefusion}.
However, SparseFusion relies on BEV feature maps to generate instances, which means it is also a dense fusion framework.

\noindent{\textbf{Sparse Fusion.}} The sparse approach, on the other hand, does not rely on dense feature maps. 
The most popular approaches are the point-level methods~\cite{pointpainting, epnet, fusionpainting}, which project LiDAR points onto the image plane to obtain image semantic cues. PointPainting~\cite{pointpainting} paints semantic scores, while PointAugmenting~\cite{pointaugmenting} paints image features. 
Also, there are some instance-level fusion~\cite{clocs,deepfusion, autoalign} methods. 
MVP~\cite{mvp} uses instance information to produce virtual points to enrich point clouds.
FrustumPointNet~\cite{frustumpointnet} and FrustumConvnet~\cite{fconvnet} utilize frustums to generate proposals. 
These methods work without relying on dense feature maps.
% \li{add a table here to help story-telling like SparseFusion}

\section{Preliminary: Fully Sparse 3D Detector}
We first briefly introduce the emerging fully sparse 3D object detector FSD~\cite{fsd}, which is adopted as our LiDAR-only baseline.
FSD is divided into two modules: Points-to-Instances and Instances-to-Boxes.

\subsection{Points-to-Instances}
% \rev{test this command \emph{use}}
The Points-to-Instances module performs 3D instance segmentation in two steps: foreground point extraction and foreground point grouping.
First, FSD classifies each point to gain the set of foreground points.
Then, the foreground points are aggregated into groups according to their spatial proximity.
A group of foreground points is denoted as a 3D instance.

\subsection{Instances-to-Boxes}
The instances-to-boxes module is divided into two parts: instance feature extraction and instance-based prediction.
First, given the instances, a PointNet-like instance feature extractor, named Sparse Instance Recognition (SIR) module is utilized for feature extraction.
Next, a simple MLP head predicts 3D bounding boxes based on extracted instance features.

The structure mentioned above does not involve any dense BEV feature maps.
Thus it can be easily extended to long-range scenarios. 

\subsection{Limitations of single-modal FSD}
\begin{figure}[h]
\centering
\includegraphics[width=1.0\columnwidth]{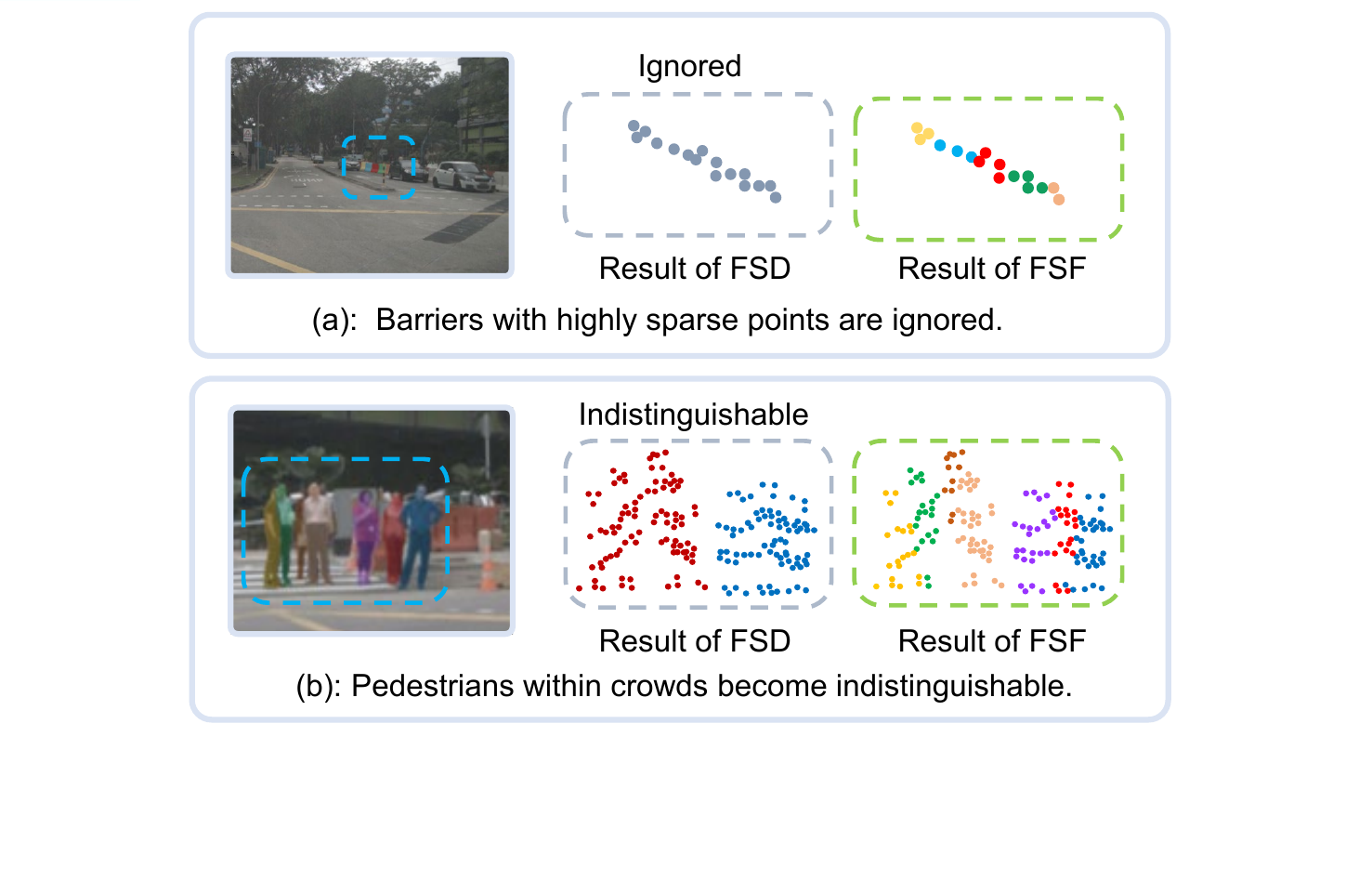}
\caption{(a): 3D instance segmentation is prone to ignore objects whose points are few. (b): It is hard for 3D segmentation to separate the overlapped objects in a crowded scene. On the contrary, handling these cases via 2D instance segmentation is easier.
}
\label{fig:inst_seg}
\end{figure}

While FSD showcases impressive capabilities, it still has several limitations.
The capabilities of FSD greatly depend on the effectiveness of 3D instance segmentation. 
However, in scenarios with highly sparse point clouds, the system might miss detecting some objects.
Additionally, when multiple objects are closely spaced, the system might mistakenly group them as one, as shown in Fig.~\ref{fig:inst_seg}.

We develop our Fully Sparse Fusion (FSF) framework, which enriches the fully sparse architecture with image information. 
Incorporating image information provides a feasible solution to these issues.

\section{Methodology}

\subsection{Overall Architecture}
\begin{figure*}[!ht]
\centering
\includegraphics[width=\textwidth]{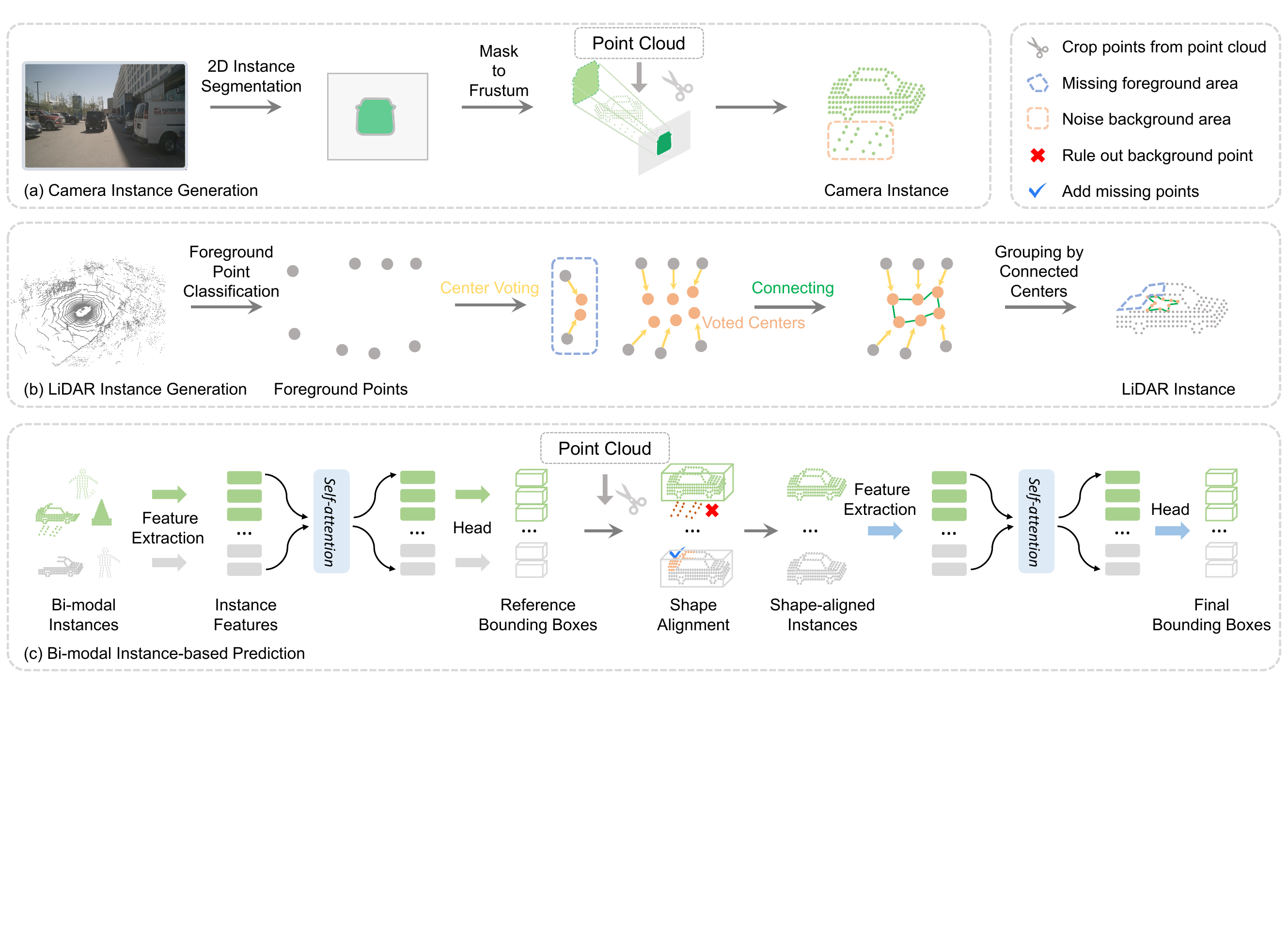}
\caption{The overview of our framework.
Our framework is mainly divided into two parts: Bi-modal Instance Generation module in \S\ref{sec:inst_generation} and Bi-modal Instance-based Prediction module in \S\ref{sec:inst_refine}. Bi-modal Instance Generation module generates instances from camera and LiDAR modalities. Then, the Bi-modal Instance-based Prediction module aligns the shapes of bi-modal instances and produces the final bounding boxes.
}
\label{fig:framework}
\end{figure*}

The overall architecture of FSF is shown in Fig.~\ref{fig:framework}. The framework is divided into two parts: the Bi-modal Instance Generation module in \S\ref{sec:inst_generation} and the Bi-modal Instance-based Prediction module in \S\ref{sec:inst_refine}.
The Bi-modal Instance Generation module produces LiDAR instances in \S\ref{sec:lidar inst generation} and camera instances in \S\ref{sec:camera inst generation}.
Then, the Bi-modal Instance-based Prediction module aligns the shape of the bi-modal instances and produces final bounding boxes based on these  instances.

\subsection{Bi-modal Instance Generation}\label{sec:inst_generation}
The Bi-modal Instance Generation module primarily consists of two parts: LiDAR Instance Generation module in \S\ref{sec:lidar inst generation} and Camera Instance Generation module in \S\ref{sec:camera inst generation}. 
The LiDAR Instance Generation uses 3D instance segmentation to generate LiDAR instances. Meanwhile, the Camera Instance Generation module uses 2D instance segmentation to create masks and then generates camera instances based on these masks.

\subsubsection{LiDAR Instance Generation}\label{sec:lidar inst generation}
Given a point cloud, we utilize 3D instance segmentation to generate a set of LiDAR instances following FSD. The detailed algorithm is introduced as follows.

First, we extract foreground points from the point cloud.
Specifically, the point cloud undergoes voxelization, converting the point cloud into a set of voxels.
A sparse voxel encoder extracts voxel features from these voxels.
Given voxel features, we obtain the feature of each point by concatenating the coordinate of this point with the corresponding voxel feature.
Using the point features, we classify each point with a simple MLP, producing the foreground score of this point.
We retain only the points with classification scores higher than $\tau$ as foreground points. 
The foreground point set is denoted as $\mathbf{F}$.

Next, we group $\mathbf{F}$ into 3D instances.
Specifically, each foreground point predicts a proxy point. 
This proxy point represents the center of the object to which this foreground point belongs. 
We call these proxy points ``voted centers''. 
The set of voted centers is denoted as $\mathbf{C}$. 
Then, the \emph{Connected Components Labeling (CCL)} algorithm~\cite{ccl} is employed to generate instances by connecting voted centers.
Within the algorithm, two voted centers are considered ``connected'' if their distance falls below a threshold \( \theta\). 
CCL groups the adjacent voted centers and forms $m^L$ groups.
The $i$-th center group is denoted as $\mathbf{C}_i$, where $i = 1, ..., m^L$. 
Given that a voted center corresponds one-to-one with a foreground point, we use $\mathbf{C}_i$ to construct the corresponding foreground point set $\mathbf{F}_i$, where each foreground point in $\mathbf{F}_i$ corresponds its voted center in $\mathbf{C}_i$.
The $i$-th LiDAR instance is $\mathbf{I}^L_i = \mathbf{F}_i$.

During training, we consider all points that fall within the ground-truth bounding box as the ground-truth foreground points. 
The centers of the ground-truth bounding boxes serve as the ground-truth voting centers for these foreground points.

\subsubsection{Camera Instance Generation}\label{sec:camera inst generation}
Camera Instance Generation module is a module parallel to LiDAR Instance Generation module for camera instance generation.
We introduce this module as follows.

First, we project the 3D LiDAR points into a 2D image plane.
Given the Intrinsic matrix $\mathbf{K}$ and Extrinsic matrix $[\mathbf{R}|\mathbf{t}]$ of a camera, the point cloud $\mathcal{P}$ are projected onto a 2D image plane to obtain the 2D points $\mathcal{U}$ by 
\begin{equation}
     \mathcal{U} = \mathbf{K}[\mathbf{R}|\mathbf{t}]\mathcal{P}.
\end{equation}
We neglect the homogeneous coordinates for simplicity.

Then, given the projected points, we utilize 2D instance segmentation to generate camera instances.
Specifically, we produce $n$ instance masks from a single-view image by the well-studied 2D instance segmentation model~\cite{htc}.
We represent the $j$-th mask as $M_j$, where $j = 1, ..., n$.
Subsequently, we collect the 2D points that fall within \( M_j \), defined as \( \mathcal{U}_j \). 
The corresponding 3D points denoted as \( \mathcal{P}_j \). 
As a result, we obtain the $j$-th camera instance as \( \mathbf{I}^C_j = \mathcal{P}_j \).
Geometrically, this process is equal to lifting a 2D mask into a 3D frustum space and collecting all the 3D points falling in this frustum.

In multi-view scenes, overlapping of frustums from different cameras occurs. 
To resolve this, we employ a point clone strategy for the regions where frustums intersect. 
Specifically, points located in the overlapping regions are cloned and allocated to each intersecting frustum. 
As a result, each frustum is guaranteed to have all relevant points.

So far, we have obtained LiDAR instances $\mathbf{I}^L = \{\mathbf{I}^L_1, \ldots, \mathbf{I}^L_{m^L}\}$ and camera instances $\mathbf{I}^C = \{\mathbf{I}^C_1, \ldots, \mathbf{I}^C_{m^C}\}$, where $m^L, m^C$ represents the number of LiDAR instances and the camera instances captured from all cameras.

\subsection{Bi-modal Instance-based Prediction}\label{sec:inst_refine}
We have generated bi-modal instances.
However, the generation methods of LiDAR instances and camera instances exhibit several limitations.
The generation of LiDAR instances is based on connecting the voted centers.
When the distance between voted centers is greater than the threshold $\theta$, some voted centers will be unconnected.
As a result, the LiDAR instances will miss a portion of foreground points. 
In contrast, camera instances are derived from frustums.
A frustum encompasses a large area and may include many noise points. 
Therefore, the shapes of LiDAR and camera instances are imprecise. Additionally, the shape distributions of the bi-modal instances diverge significantly.
These issues pose difficulties in effective feature learning, which is critical for achieving high-quality detection.

To address these issues, we propose a Bi-modal Instance-based Prediction module. 
This module aligns the shapes of bi-modal instances and then outputs the final prediction results based on the shape-aligned instances.
It contains four parts: instance feature extraction, instance feature interaction, instance shape alignment, and the final prediction head.

\subsubsection{Instance Feature Extraction}
% First, we combine the instances from different modalities as $\mathbf{I} = \mathbf{I}^L \cup \mathbf{I}^C$. 
% We extract instance features from $\mathbf{I}$.
The initial step is extracting features from bi-modal instances to facilitate downstream processing tasks.  
We apply Sparse Instance Recognition (SIR) module~\cite{fsd} in FSD, a PointNet-like module, on instances to extract the instance features.
Since LiDAR instances and camera instances have totally different shapes, we use two separate SIR modules to extract the features of LiDAR instances and camera instances.
Specifically, we use $S$ to denote the modality, where $S$ can be either $L$ or $C$, denoting LiDAR or camera.
For the $S$ modality, we denote \( k \)-th instance as \( \mathbf{I}^S_k \), $k = 1, \ldots, m^S$, where $m^S$ represents the total number of instances in modality $S$. 
Assume \( \mathbf{I}^S_k \) comprises \( h \) points. That is, \( \mathbf{I}^S_k = \{\mathbf{p}^k_1, \mathbf{p}^k_2, \ldots, \mathbf{p}^k_h\} \). 
The corresponding  point feature is denoted as $\mathcal{G}^S_k = \{\mathbf{g}^k_1, \mathbf{g}^k_2, \ldots, \mathbf{g}^k_h\}$.
Then, the feature extraction is formulated as
\begin{equation}
    \mathbf{f}^S_k = \texttt{SIR}(\mathcal{G}^S_k),
\end{equation}
where the instance feature $\mathbf{f}^S_k \in \mathbb{R}^{D}$ is a single feature vector. $D$ represents the dimension of the feature vector.

\subsubsection{Instance Feature Interaction}\label{sec:inst_feat_interaction}
Given the instance features, we propose an Instance Feature Interaction module to leverage the complementarity of LiDAR and camera modalities.
This module facilitates interaction among instances by applying self-attention to all instance features. 
Specifically, given the extracted LiDAR instance features $\mathbf{f}^L = \{\mathbf{f}^L_1, \ldots, \mathbf{f}^L_{m^L}\}$ and camera instance features $\mathbf{f}^C = \{\mathbf{f}^C_1, \ldots, \mathbf{f}^C_{m^C}\}$, 
we combine them into bi-modal instance features as $\mathbf{f} = \mathbf{f}^L \cup \mathbf{f}^C$.
$m^L$ represents the number of LiDAR instances and $m^C$ represents the number of camera instances.
Next, we have
\begin{equation}
    \mathbf{f}' = \texttt{SelfAttn}(\mathbf{f}),
\end{equation}
where \( \mathbf{f}' = \{ \mathbf{f}'_1, \ldots, \mathbf{f}'_m \} \) represents the updated instance features.
$m = m^L + m^C$ represents the number of instances.

\subsubsection{Instance Shape Alignment}\label{sec:inst_shape_refine}
As mentioned earlier, the bi-modal instances have totally different shape distributions, which is adverse to performing high-quality bi-modal detection.
Therefore, we introduce an Instance Shape Alignment module to produce shape-aligned instances.
Specifically, 
we divide  \( \mathbf{f}'\) into $\mathbf{f}'_L$ and $\mathbf{f}'_C$.
Then, we use two separate MLP heads to predict \textit{reference} boxes for bi-modal instances.
Then, by placing a predicted reference box into the 3D point cloud space, we include points that fall within this box and exclude those points that are outside of the box.
As a result, we obtain the shape-aligned instance of this box.

% By instance shape alignment, we produce the refined instances that share the same shape distribution.
% As a result, the bounding boxes generated by these refined instances are much better than those generated by the initial bi-modal instances.

\subsubsection{Final Prediction Head}\label{sec:inst_pred}
Compared with the initial bi-modal instances, the instances that are aligned in shapes exhibit enhanced precision and more consistent distribution of their shapes. 
This improvement makes them suitable for the prediction of final bounding boxes.
Specifically, another SIR module, self-attention module, and MLP head are utilized to produce the \textit{final} bounding boxes based on these shape-aligned instances.
The bounding box is defined by its central coordinates, dimensions, and orientation.

\subsection{Bi-modal Instance Label Assignment}\label{sec:query_assign}
% During training, it is essential to assign correct labels to the bi-modal instances. 
% However, each predicted instance inevitably incorporates noise points, especially before the shape alignment. 
Assigning an instance to the correct ground truth bounding box is challenging due to these noise points.
In this section, we aim to design an effective assignment strategy for bi-modal instances.
Before introducing our assignment strategy, we first take a quick overview of the assignment strategy used in LiDAR-only FSD.
FSD regards the instance whose center falls into GT boxes as positive, where the center of an instance is defined as the average of all the points in this instance.
We name this strategy \emph{point-in-box assignment}.
This works for most LiDAR instances.

\begin{figure}[h]
\centering
\includegraphics[width=1.0\columnwidth]{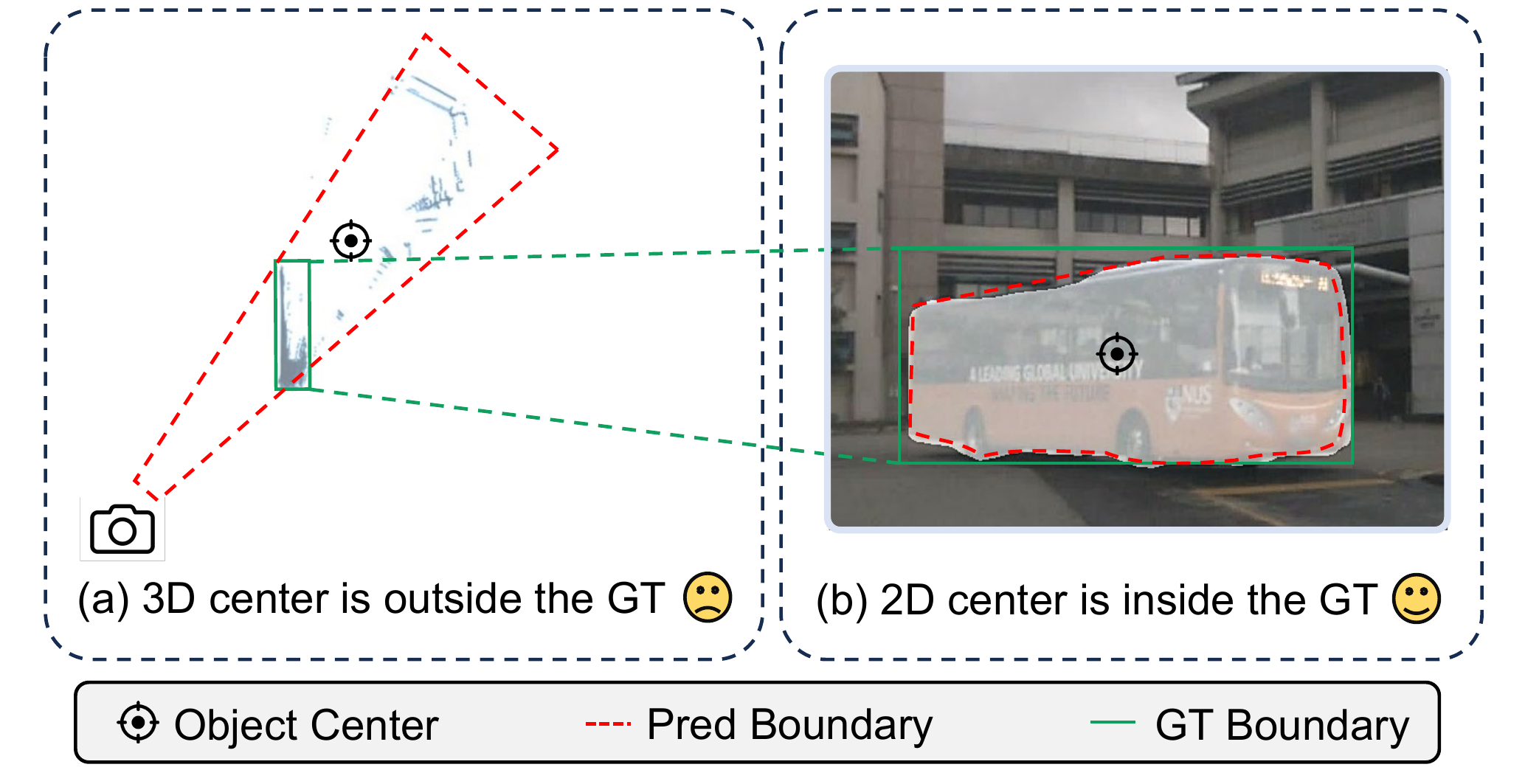}
\caption{The motivation of two-stage assignment. 
The center of a camera instance is hard to fall into 3D GT boxes due to noise points as (a) shows. 
However, it is easy to assign this camera instance to the corresponding GT on the 2D plane as (b) demonstrates.
}
\label{fig:assign}
\end{figure}

However, this strategy could hardly work for camera instances. 
A camera instance usually has many noise points.
The noise points are divided into two types: noise background points and noise foreground points.
(i) Camera instances usually contain a large number of irrelevant background points because the frustums are extended along the depth direction and cover a large area as Fig.~\ref{fig:assign} shows. 
(ii) The mask generated for an object may inadvertently encompass segments of an adjacent object, especially in cases where two objects are in a direct line with the camera's viewpoint but are located at different distances from the camera. 
This introduces noise into the camera instance, manifesting as notable depth disparities within foreground points.

% \begin{itemize}
%     \item Camera instances usually contain a large amount of irrelevant background points because the frustums are extended along the depth direction and cover a large area as Fig.~\ref{fig:assign} shows. 
%     These background points lead to a displacement of the central point.
%     This makes the center of a camera instance easily fall outside its corresponding ground truth bounding box.
    
%     \item Due to the potential error of 2D instance segmentation, a single frustum may contain points from multiple foreground objects.
%     For example, the mask \( M_a \) of a partially occluded object \( O_a \) might encompass portions of the occluding object \( O_b \). 
%     Given the depth difference between \( O_a \) and \( O_b \), this overlap also introduces a displacement of the central point.
% \end{itemize}

These noise points, which are challenging to eliminate, result in displacements of the centers. Such displacements can cause the centers of camera instances to deviate outside their corresponding ground truth bounding boxes. 
As a result, a substantial number of camera instances are erroneously assigned as negative during the point-in-box assignment process.

To address this issue, we propose a Bi-modal instance assignment as Algo.~\ref{algo:tow_round_assignment} shows.
Briefly, our assignment strategy is divided into two stages: the first stage is a LIDAR-based assignment stage, which operates in 3D space and is an enhanced version of the point-in-box assignment strategy. 
The second stage is a camera-based assignment stage, which operates in 2D space and assigns instances based on the 2D similarity between the instances and the projected ground truths.

\subsubsection{Stage 1: LiDAR-based Assignment}\label{sec:lidar_based_assign}
In FSD, the point-in-box strategy simply averages the coordinates of all points in an instance to obtain the center of this instance. 
In order to eliminate the influence of irrelevant background points, we introduce a weighted averaging approach.
Specifically, the $k$-th bi-modal instance is denoted as \( \mathbf{I}_k = \{\mathbf{p}^k_1, \ldots, \mathbf{p}^k_l, \ldots, \mathbf{p}^k_h\} \).
$h$ represents the number of points in $\mathbf{I}_k$ and $l = 1, \ldots, h$ represents the index of the point.
In \S\ref{sec:lidar inst generation}, we produce the foreground classification scores for all the points in the point cloud.
Therefore, for $\mathbf{p}^k_l$, we have the corresponding classification score $s^k_l$.
With the help of $s^k_l$, we define the weighted center of $\mathbf{I}_k$ as
\begin{equation}\label{eq:weighted_center}
\texttt{Center}(\mathbf{I}_k) = \frac{\sum_{l=1}^{h} s^k_l \cdot \mathbf{p}^k_l}{\sum_{l=1}^{h} s^k_l}.
\end{equation}
Then, we use $\texttt{Center}(\mathbf{I}_k)$ as the center of $\mathbf{I}_k$ to perform point-in-box assignment. 
We call the whole process above LiADR-based assignment.

\begin{algorithm}[t]
\caption{Bi-modal Instance Assignment Strategy}
\begin{algorithmic}[1]
\label{algo:tow_round_assignment}
\renewcommand{\algorithmicrequire}{\textbf{Input:}}
\renewcommand{\algorithmicensure}{\textbf{Output:}}
\REQUIRE \ \\
Predicted 3D Instances: $\mathbf{I}$, \\
GT bounding boxes $\mathbf{B}^*$.
\ENSURE \  \\
Assigned labels $\mathbf{L}$.
\\ \hrulefill

\STATE $L_i = \textit{NaN}$ for all $L_i$ in $\mathbf{L}$ \algorithmiccomment{Initialize to NaN.}
\vspace{2mm}
\STATE \textit{Stage-1: LiDAR-based Assignment}
\FOR{$\mathbf{I}_i$ in $\mathbf{I}$}
    \STATE $c_i$ = \texttt{Center}($\mathbf{I}_i$) .\algorithmiccomment{$\texttt{Center}$ is defined in \S\ref{sec:lidar_based_assign}.}
    \FOR{$\mathbf{B}^*_j$ in $\mathbf{B}^*$}
        \IF{$c_i$ falls inside $\mathbf{B}^*_j$}
            \STATE Assign $\mathbf{B}^*_j$ to $\mathbf{I}_i$, i.e., $L_i = j$.
            \STATE \textbf{break}
        \ENDIF
    \ENDFOR
\ENDFOR

\vspace{3mm}
\STATE \textit{Stage-2: Camera-based Assignment}
\STATE \textit{Define} $\mathcal{O}$, such that $\mathcal{O}(\mathbf{I}_i) = L_i$.
\STATE $\mathbf{I} \leftarrow \{\mathbf{I}_i | \mathcal{O}(\mathbf{I}_i)~is~\textit{NaN}\}$. \algorithmiccomment{Select instances that are still unassigned after Stage-1.}
\vspace{-3mm}
\FOR{$\mathbf{I}_i$ in $\mathbf{I}$} 
    \STATE Initialize $\alpha$. \algorithmiccomment{Initialize the threshold.}
    \FOR{$\mathbf{B}^*_j$ in $\mathbf{B}^*$}
        \STATE $\beta = \mathcal{S}(\mathbf{I}_i, \mathbf{B}^*_j)$.\algorithmiccomment{2D similarity $\mathcal{S}$ is defined in \S\ref{sec:cam_based_assign}.}
        \vspace{-3mm}
        \IF{$\beta > \alpha$}
            \STATE $\alpha = \beta$.\algorithmiccomment{Update the threshold.}
            \STATE $L_i = j$.
        \ENDIF
    \ENDFOR
\ENDFOR

\RETURN $\mathbf{L}$
\end{algorithmic}
\end{algorithm}
\subsubsection{Stage 2: Camera-based Assignment}\label{sec:cam_based_assign}
Although Eq.~\ref{eq:weighted_center} refines the centers, the center of a camera instance deviates outside the GT box still happens. 
There are two reasons: 
(i) The predicted classification scores may be incorrect. 
(ii) Eq.~\ref{eq:weighted_center} does not resolve the issues with the notable depth disparities within the foreground points.
To rectify this, we implement a second assignment stage named the camera-based assignment stage. 
To avoid conflicts between the LiDAR-based assignment and the camera-based assignment, the second stage only serves for the camera instances that are still unassigned after the first stage.

The second assignment stage assigns labels by 2D similarity.
Specifically, the 2D instance segmentation joint produces 2D masks and 2D bounding boxes of the foreground objects.
Therefore, for the $j$-th remaining unassigned camera instance $\mathbf{I}^C_j$, we have a corresponding 2D bounding box $\mathbf{V}_j$.
Besides, we can project the $d$-th 3D ground-truth box $\mathbf{B}^*_d$ onto 2D image plane to obtain the 2D box $\mathbf{V}^{*}_d$.
As a result, we can compute the 2D similarity $\mathcal{S}$ between $\mathbf{I}^C_j$ and $\mathbf{B}^*_d$ by 
\begin{equation}
\mathcal{S}(\mathbf{I}^C_j, \mathbf{B}^*_d) = \text{IoU}_{2D}(\textbf{V}_j, \textbf{V}^{*}_d).
\end{equation}
where $\text{IoU}_{2D}$ means 2D Intersection over Union.

Given $\mathcal{S}(\mathbf{I}^C_j, \mathbf{B}^*_d)$, we then follow the commonly-used ``max IoU''~\cite{fasterrcnn, ssd} assignment strategy to assign labels.
For camera instance $\mathbf{I}^C_j$, ``max IoU'' strategy aims to find the 3D GT box $\mathbf{B}^*_e$, $e$ is the index, where
$\mathbf{B}^*_e$ has the max $\mathcal{S}(\mathbf{I}^C_j, \mathbf{B}^*_e)$ value among all the GT boxes.
If the maximum value $\mathcal{S}(\mathbf{I}^C_j, \mathbf{B}^*_e)$ is greater than the threshold $\alpha$, we assign $\mathbf{I}^C_j$ to $\mathbf{B}^*_e$ as lines 11-17 of Algo.~\ref{algo:tow_round_assignment} shows.

After these two stages, those instances still not associated with any GTs are assigned as negative.
This two-stage assignment strategy is used in both \S\ref{sec:inst_shape_refine} for generating reference boxes and in \S\ref{sec:inst_pred} for generating final boxes.

% In summary, the assignment strategy unfolds in two stages. 
% The first stage is LiDAR-based assignment.
% The proposed LiDAR-based assignment method demonstrates improved robustness over the original point-in-box approach in mitigating the influence of background point disturbances within an instance.
% Following this, the camera-based assignment further ensures each instance will be properly assigned by assigning instances in 2D space.

\subsection{Loss}\label{sec:head}
Our framework comprises four heads: 
a head is dedicated to classifying and voting points in \S\ref{sec:lidar inst generation}.
Two heads are dedicated to generating LiDAR reference bounding boxes and camera reference bounding boxes in \S\ref{sec:inst_shape_refine}.
And one head is used for generating final bounding boxes in \S\ref{sec:inst_pred}.

The head used for producing bounding boxes is divided into two branches, namely the regression branch and the classification branch.
The regression branches regress the 3D bounding box and velocity for each instance. 
The classification branches are used to classify instances.
We use L1 loss for regression branches and Focal Loss~\cite{focalloss} for classification branches, respectively.

Formally, the loss of the first head is denoted as $\mathcal{L}_{P} = \mathcal{L}_{P}^{cls} + \mathcal{L}_{P}^{vote}$,
where $\mathcal{L}_{P}^{cls}$ and $\mathcal{L}_{P}^{vote}$ is the point classification loss and point voting loss in FSD~\cite{fsd}.
The loss of LiDAR reference box head is $\mathcal{L}_{L} =  \mathcal{L}_{L}^{cls} + \mathcal{L}_{L}^{reg}$, where $\mathcal{L}_{L}^{cls}$ and $\mathcal{L}_{L}^{reg}$ is the classification and regression loss for LiDAR instances. 
The loss of camera reference box head is $\mathcal{L}_{C} = \mathcal{L}_{C}^{cls} + \mathcal{L}_{C}^{reg}$.
The loss of final box head is $\mathcal{L}_{F} = \mathcal{L}_{F}^{cls} + \mathcal{L}_{F}^{reg}$.

In total, we have
\begin{equation}
\begin{split}
\mathcal{L}_{total} = \mathcal{L}_{P} +  \mathcal{L}_{L} + \mathcal{L}_{C} + \mathcal{L}_{F},
\end{split}
\end{equation}
where the loss weight of each item is the same.

\section{Experiments}
\subsection{Setup}
\subsubsection{Dataset: nuScenes} Our experiment is mainly conducted on the nuScenes dataset~\cite{nus}, which provides a total of 1000 scenes.
Each scene is about 20 seconds long and contains collected data from a 32-beam LiDAR along with six cameras.
The effective detection area of nuScenes covers a $100m \times 100m$.
The annotations are divided into 10 categories.
The official metrics of nuScenes is mean Average Precision (mAP) and nuScenes Detection Score (NDS).
The NDS is calculated based on mAP, mATE (translation error), mASE (scale error), mAOE (orientation error), mAVE (velocity error), and mAAE (attribute error).
\begin{table*}[htp]
\begin{center}

\resizebox{0.90\textwidth}{!}{%
\begin{tabular}{l|c|cc|ccccccc}
  \specialrule{1pt}{0pt}{1pt}
\toprule
\multirow{2}{*}{Method} & \multirow{2}{*}{Modality} & \multicolumn{2}{c|}{\textit{val}} & \multicolumn{7}{c}{\textit{test}}                                        \\ 
% \cmidrule(l){3-11} 
                        &                           & mAP~$\uparrow$   &\cellcolor{gray!15}  NDS~$\uparrow$  & mAP~$\uparrow$ & \cellcolor{gray!15}  NDS~$\uparrow$ & mATE~$\downarrow$  & mASE~$\downarrow$   & mAOE~$\downarrow$   & mAVE$~\downarrow$  & mAAE$~\downarrow$   \\ \midrule
DETR3D~\cite{detr3d}                  & C                         & 34.9        & \cellcolor{gray!15}43.4       & 41.2       & \cellcolor{gray!15}47.9       & 0.641 & 0.255 & 0.394 & 0.845 & 0.133 \\
BEVDet4D~\cite{bevdet4d}                & C                         & 42.1        & \cellcolor{gray!15}54.5       & 45.1       &\cellcolor{gray!15}56.9       & 0.511 & 0.241 & 0.386 & 0.301 & 0.121 \\
BEVFormer~\cite{bevformer}               & C                         & 41.6        & \cellcolor{gray!15}51.7       & 48.1       & \cellcolor{gray!15}56.9       & 0.582 & 0.256 & 0.375 & 0.378 & 0.126 \\ \midrule
SECOND~\cite{second}                  & L                         & 52.6        & \cellcolor{gray!15}63.0       & 52.8       & \cellcolor{gray!15}63.3       & -     & -     & -     & -     & -     \\
CenterPoint~\cite{centerpoint}             & L                         & 59.6        &\cellcolor{gray!15}66.8       & 60.3       &\cellcolor{gray!15}67.3       & 0.262 & 0.239 & 0.361 & 0.288 & 0.136 \\
% Focals Conv~\cite{centerpoint}             & L                         & 61.2        &\cellcolor{gray!15}68.1       & 63.8       &\cellcolor{gray!15}70.0       & 0.262 & 0.239 & 0.361 & 0.288 & 0.136 \\
FSD$^*$~\cite{fsd}                  & L                         & 62.5        &\cellcolor{gray!15}68.7       & -          & \cellcolor{gray!15}-          & -     & -     & -     & -     & -     \\
VoxelNeXt~\cite{voxelnext}                  & L                         & 63.5        &\cellcolor{gray!15}68.7       & 64.5          & \cellcolor{gray!15}70.0          & 0.268     & 0.238     & 0.377     & 0.219     & 0.127     \\
TransFusion-L~\cite{transfusion}           & L                         & 64.9        &\cellcolor{gray!15}69.9       & 65.5       & \cellcolor{gray!15}70.2       & 0.256 & 0.240 & 0.351 & 0.278 & 0.129 \\
LargeKernel3D~\cite{largekernel}           & L                         & 63.3        &\cellcolor{gray!15}69.1       & 65.3       & \cellcolor{gray!15}70.5       & 0.261 & 0.236 & 0.319 & 0.268 & 0.133 \\ \midrule
% PointPainting           & C+L                       & -           & -          &\cellcolor{gray!15} 46.4       & 58.1       & 0.388 & 0.271 & 0.496 & 0.247 & 0.111 \\
FUTR3D~\cite{futr3d}                  & C+L                       & 64.5        &\cellcolor{gray!15}68.3       & -          &\cellcolor{gray!15}-          & -     & -     & -     & -     & -     \\
MVP~\cite{mvp}                     & C+L                       & 67.1        &\cellcolor{gray!15}70.8       & 66.4       & \cellcolor{gray!15}70.5       & 0.263 & 0.238 & 0.321 & 0.313 & 0.134 \\
PointAugmenting~\cite{pointaugmenting}         & C+L                       & -           &\cellcolor{gray!15}-          & 66.8       & \cellcolor{gray!15}71.0       & 0.254 & 0.236 & 0.362 & 0.266 & 0.123 \\
UVTR~\cite{uvtr}        & C+L                       & 65.4           &\cellcolor{gray!15}70.2          & 67.1       & \cellcolor{gray!15}71.1       & 0.306 & 0.245 & 0.351 & 0.225 & 0.124 \\
TransFusion~\cite{transfusion}             & C+L                       & 67.5        & \cellcolor{gray!15}71.3       & 68.9       & \cellcolor{gray!15}71.6       & 0.259 & 0.243 & 0.359 & 0.288 & 0.127 \\
BEVFusion~\cite{bevfusion-damo}               & C+L                       & 67.9        &\cellcolor{gray!15}71.0       & 69.2       & \cellcolor{gray!15}71.8       & -     & -     & -     & -     & -     \\
BEVFusion~\cite{bevfusion-mit}               & C+L                       & 68.5        & \cellcolor{gray!15}71.4       & 70.2       & \cellcolor{gray!15}72.9       & 0.261 & 0.239 & 0.329 & 0.260 & 0.134 \\
DeepInteraction~\cite{deepinteraction}         & C+L                       & 69.9        & \cellcolor{gray!15}72.6       & \textbf{70.8}      & \cellcolor{gray!15}73.4       & 0.257 & 0.240 & 0.325 & 0.245 & 0.128 \\
FSF (Ours)              & C+L                       &    \textbf{70.4}      &\cellcolor{gray!15}\textbf{72.7}      &70.6       & \cellcolor{gray!15}\textbf{74.0}       & \textbf{0.246} & \textbf{0.234} & \textbf{0.318} & \textbf{0.211} & \textbf{0.123} \\ 
\bottomrule
  \specialrule{1pt}{1pt}{0pt}
\end{tabular}%
}
\end{center}
\caption{
Comparison with state-of-the-art methods on the nuScenes dataset. 
We do not use test-time augmentation or model ensemble.
$^*$: Reimplemented by us. 
We mark the official benchmark of nuScenes as \colorbox{gray!15}{Gray}.
C: camera modality.
L: LiDAR modality.
}
\label{tab:sota_compare}
\end{table*}

\subsubsection{Dataset: Waymo Open Dataset (WOD)}
To further showcase our approach, we test our method on the widely-used Waymo Open Dataset (WOD)~\cite{wod}. The WOD has 1150 sequences, broken down into 798 for training, 202 for validation, and 150 for testing. 
It totally contains more than 200k frames.
WOD is equipped with a 64-beam LiDAR and 5 cameras. 
However, the cameras do not cover a full 360-degree view.
There is a blind area at the back.
Objects in the WOD can be detected up to 75 meters away, covering an area sized $150m \times 150m$.
For metrics, WOD employs mAP and mAPH, where mAPH represents mean average precision weighted by heading.
In WOD, objects are categorized into LEVEL 1 or LEVEL 2.
An object is marked as LEVEL 2 if the number of points in this object is less than or equal to 5.
The other objects are marked as LEVEL 1.

\subsubsection{Dataset: Argoverse 2}
To demonstrate our superiority in long-range detection, we proceed to conduct experiments on the Argoverse 2 dataset~\cite{argo2}, abbreviated as AV2.
The AV2 contains 150k annotated frames, 5$\times$ larger than nuScenes.
AV2 employs two 32-beam LiDARs to form a 64-beam LiDAR, combined with 7 surrounding cameras.
The valid detection distance of AV2 is 200m (covering $400m \times 400m$ area).
In addition, AV2 contains 30 classes, exhibiting challenging long-tail distribution.
For metrics, in addition to the mean Average Precision (mAP), AV2 provides a Composite Detection Score (CDS) benchmark. CDS is the average score based on mAP, mATE, mASE, and mAOE.

\subsubsection{Model}
Following FSD~\cite{fsd}, we adopt a SparseUNet~\cite{parta2} as the LiDAR backbone. 
The voxel size is set to $[0.2m, 0.2m, 0.2m]$. 
With respect to 2D instance segmentation, we use HybridTaskCascade (HTC)~\cite{htc} pre-trained on nuImages~\cite{nus}.  
For the SIR module that utilized for instance feature extraction, we use the same setting as FSD~\cite{fsd}.
For the thresholds, we set the threshold $\tau$ of the foreground classification to $0.1$. 
The threshold $\theta$ of CCL is $0.2$ and the threshold $\alpha$ of the ``max IoU'' strategy is $0.3$. 

\subsubsection{Training Scheme} 
For nuScenes, consistent with previous methods\cite{transfusion, bevfusion-damo}, we first train the LiDAR-only detector.
We pre-train FSD on nuScenes for 20 epochs.
Next, initializing from the pre-trained FSD, we train our model for 6 epochs with CBGS~\cite{cbgs}.
For the augmentation, following FSD~\cite{fsd}, we randomly rotate, scale, and translate the point cloud.
We also randomly shuffle the order of points.
The optimizer is AdamW~\cite{adamw}.
We adopt the one-cycle learning rate policy, with a maximum learning rate of $10^{-4}$ and a weight decay of $10^{-2}$. The model is trained on eight NVIDIA RTX 3090 with a batch size of 8.
For AV2, we pre-train FSD for 12 epochs and continue training FSF for another 6 epochs. 
We adopt the same data augmentation as FSD. 
We set the initial learning rate as $10^{-4}$ with a weight decay of $0.05$.
For Waymo Open Datset~\cite{wod}, FSD is pre-trained for 6 epochs, followed by 6 epochs for our model. 
The data augmentation is the same as FSD.
We use the learning rate of $3\times10^{-5}$ with a weight decay of $0.05$.
Also, we set the weight decay of the batch norm to $0$ and set the batch size as 16 following FSD.

\subsection{Comparison with State-of-the-art Methods}
% \vspace{-3mm}
\subsubsection{Results on nuScenes}
As illustrated in Table~\ref{tab:sota_compare}, our proposed method, FSF, outperforms all preceding multi-modal 3D object detection techniques on the official nuScenes detection score (NDS) benchmark. 
It is particularly remarkable that FSF exhibits superior efficacy in comparison to the leading-edge methods such as BEVFusion~\cite{bevfusion-damo, bevfusion-mit} and DeepInteraction~\cite{deepinteraction}. 
Moreover, our method achieves SOTA performance on the five detailed metrics: mean Average Translation Error (mATE), mean Average Scaling Error (mASE), mean Average Orientation Error (mAOE), mean Average Velocity Error (mAVE) and mean Average Attribute Error (mAAE), which further demonstrates the superiority of our approach.

\subsubsection{Results on Waymo Open Dataset (WOD)}
\begin{table}[h]
\begin{center}
\resizebox{0.99\linewidth}{!}{
\begin{tabular}{l|c|c|c|c|c}
\toprule
Methods & Modality & \textit{Overall} & \textit{Vehicle} & \textit{Pedestrian} & \textit{Cyclist} \\
\midrule
PointPillars~\cite{pointpillar} & L & 57.8 & 63.1 & 50.3 & 59.9 \\
LiDAR-RCNN~\cite{lidarrcnn} & L & 61.3 & 67.9 & 51.7 & 64.4 \\
3D-MAN++$\dag$~\cite{3dman} & L & 63.0 & - & - & -  \\
PVRCNN~\cite{pvrcnn} & L & 63.3 & 68.4 & 57.6 & 64.0 \\
TransFusion-L$\ddag$~\cite{transfusion} & L & 64.9 & 65.1 & 63.7 & 65.9  \\
CenterPoint~\cite{centerpoint} & L & 67.6 & 68.4 & 65.8 & 68.5 \\
FSD$^\ast$~\cite{fsd} & L & 69.7 & 68.5 & 68.0 & 72.5 \\
\midrule
PointAugmenting~\cite{pointaugmenting} & C+L & 66.7  & 62.2 & 64.6 & 73.3 \\
TransFusion~\cite{transfusion} & C+L & 65.5 & 65.1 & 64.0 & 67.4 \\
DeepFusion~\cite{deepfusion} & C+L & 67.0 & - & - &  \\
LoGoNet~\cite{logonet} & C+L & 71.4 & 70.7 & 69.9 & 73.5\\
FSF & C+L & \textbf{72.3} & 70.0 & 70.4 & 76.5 \\
\bottomrule
\end{tabular}}
\end{center}
\caption{Performance on the Waymo validation set measured by LEVEL 2 mAPH. All the methods are under the single-frame setting for fair comparison. $\dag$: Baseline of DeepFusion, $\ddag$: Baseline of TransFusion  $^\ast$: Baseline of ours.}
\label{tab:waymo_level2}
\end{table}

Table~\ref{tab:waymo_level2} compares cutting-edge 3D object detection methods on the Waymo validation set using the LEVEL 2 mAPH metric. 
Our method FSF outperforms others, achieving the SOTA performance. 
Specifically, we achieve the SOTA performance of 70.4 mAPH in Pedestrians, and 76.5 mAPH in Cyclists. 
In summary, FSF achieves a leading performance of 72.3 mAPH,  highlighting its excellence in multi-modal 3D object detection.

\setlength{\tabcolsep}{3pt}
\begin{table*}[!ht]
% \vspace{-2mm}
\begin{center}
\resizebox{\textwidth}{!}{%
\begin{tabular}{l|l|ccccccccccccccccccccccccc}
\specialrule{1pt}{0pt}{1pt}
\toprule
    &
  \textbf{Methods} &
 \rotatebox{90} {Average}& 
 \rotatebox{90} {Vehicle} & 
 \rotatebox{90} {Bus} &
 \rotatebox{90} {Pedestrian} &
 \rotatebox{90} {Box Truck} &
 \rotatebox{90} {C-Barrel} &
 \rotatebox{90} {Motorcyclist} &
%  \rotatebox{90} {Truck} &
 \rotatebox{90} {MPC-Sign} &
 \rotatebox{90} {Motorcycle} &
 \rotatebox{90} {Bicycle} &
 \rotatebox{90} {A-Bus} &
 \rotatebox{90} {School Bus} &
 \rotatebox{90} {Truck Cab} &
 \rotatebox{90} {C-Cone} &
 \rotatebox{90} {V-Trailer} &
 \rotatebox{90} {Bollard} &
 \rotatebox{90} {Sign} &
 \rotatebox{90} {Large Vehicle} &
 \rotatebox{90} {Stop Sign} &
 \rotatebox{90} {Stroller} &
 \rotatebox{90} {Bicyclist} &
\\ 
 % \midrule
% \textit{mAP} &&&& \\
\midrule
\multirow{4}{*}{mAP} &
CenterPoint\ddag~\cite{centerpoint}        & 13.5 & 61.0 & 36.0  & 33.0  & 26.0    & 22.5  & 16.0   & 16.0 & 12.5 & 9.5  & 8.5 & 7.5 & 8.0 & 8.0  & 7.0 & 25.0 & 6.5  &3.0 & 28.0 & 2.0 & 14  \\
&CenterPoint$^\ast$~\cite{centerpoint}        & 22.0 & 67.6 & 38.9 & 46.5 & 40.1 & 32.2 & 28.6 & 27.4 & 33.4 & 24.5 & 8.7 & 25.8 & 22.6 & 29.5  & 22.4 & 37.4 & 6.3 & 3.9 & 16.9 & 0.5 & 20.1\\
&VoxelNeXt~\cite{voxelnext}        & 30.5 & 72.0 & 39.7 & 63.2 & 39.7 & 64.5 & 46.0 & 34.8 & 44.9 & 40.7 & 21.0 & 27.0 & 18.4 & 44.5  & 22.2 & 53.7 & 15.6 & 7.3 & 40.1 & 11.1 & 34.9\\
&FSD~\cite{fsd} (Baseline)       & 28.2 & 68.1 & 40.9 & 59.0 & 38.5 & 42.6 & 39.7 & 26.2 & 49.0 & 38.6 & 20.4 & 30.5 & 14.8 & 41.2  & 26.9 & 41.8 & 11.9 & 5.9 & 29.0 & 13.8 & 33.4\\
&FSF  & \textbf{33.2} & 70.8 & 44.1 & 60.8 & 40.2  & 50.9 & 48.9 & 28.3 & 60.9 & 47.6 & 22.7 & 36.1 & 26.7 & 51.7  & 28.1 & 41.1 & 12.2 & 6.8 & 27.7 & 25.0 & 41.6\\
% \midrule
% \textit{CDS} &&&& \\
\midrule
\multirow{3}{*}{CDS} &
CenterPoint$^\ast$~\cite{centerpoint}        & 17.6  & 57.2  & 32.0  & 35.7   & 31.0   & 25.6  & 22.2 & 19.1 & 28.2  & 19.6 & 6.8  & 22.5 & 17.4  & 22.4 & 17.2 & 28.9 & 4.8 & 3.0 & 13.2 & 0.4 & 16.7  \\
&VoxelNeXt~\cite{voxelnext}                & 23.0 & 57.7   & 30.3  & 45.5 & 31.6   & 50.5  & 33.8 & 25.1 & 34.3  & 30.5
& 15.5 & 22.2 & 13.6  & 32.5 & 15.1 & 38.4 & 11.8 & 5.2 & 30.0 & 8.9 & 25.7 \\
&FSD~\cite{fsd} (Baseline)           & 22.7 & 57.7   & 34.2  & 47.5 & 31.7   & 34.4  & 32.3 & 18.0 & 41.4  & 32.0 & 15.9 & 26.1 & 11.0  & 30.7 & 20.5 & 30.9 & 9.5 & 4.4 & 23.4 & 11.5 & 28.0 \\
&FSF          & \textbf{25.5} & 59.6   & 35.6  & 48.5  & 32.1  & 40.1  & 35.9 & 19.1 & 48.9  & 37.2 & 17.2 & 29.5 & 19.6  & 37.3 & 21.0 & 29.9 & 9.2 & 4.9 & 21.8 & 18.5 & 32.0 \\
\bottomrule
\specialrule{1pt}{1pt}{0pt}
\end{tabular}%
}
\end{center}
\caption{
Comparison with state-of-the-art methods on Argoverse 2 validation split.
C-Barrel: construction barrel.
MPC-Sign: mobile pedestrian crossing sign.
A-Bus: articulated bus.
C-Cone: construction cone.
V-Trailer: vehicular trailer.
\ddag: provided by authors of AV2 dataset.
$^\ast$: reimplemented by FSD.
Some categories are excluded from the table due to the limited number of instances they contain. 
However, the average results consider all categories, even those that are omitted.
}
\label{tab:sota_argo}
\end{table*}

\subsubsection{Results on Argoverse 2}
AV2 has a much larger perception range than nuScenes, making it a suitable testbed to demonstrate our superiority in long-range detection.
The results in Table~\ref{tab:sota_argo} show that FSF consistently outperforms previous LiDAR-based state-of-the-art methods by large margins. 
Notably, we observe an 8 to 10 mAP increase in categories with relatively small sizes, such as Traffic Cone, Motorcycle, and Bicycle. 
Although previous multi-modal methods like BEVFusion~\cite{bevfusion-damo} and TransFusion~\cite{transfusion} closely approach our results on the nuScenes dataset, the computational burden of their dense BEV feature maps proves to be prohibitive for the AV2 dataset.
Specifically, the size of detection area of nuScenes is $100m \times 100m$ while the size of AV2 is $400m \times 400m$.
On AV2, the BEV-feature-map-based methods require 16 times as much memory as nuScenes for BEV-map-related operations.
As a result, we can not reproduce them on AV2 due to the limitation of GPU memories.

\subsection{Ablation Studies}
In this section, we analyze each module within \name to understand the significance of each component.

\subsubsection{Bi-modal Instances Generation}
\begin{table*}[!htb]
\begin{center}
\resizebox{0.9\textwidth}{!}{
\setlength{\tabcolsep}{5pt}
\begin{tabular}{ccc|cc|cccccccccc}
\toprule
 \makecell{Cam. \\ Instance} &  \makecell{LiDAR \\ Instance} & \makecell{Shpae \\ Align.} & NDS  & mAP  & Car  & Truck & Bus  & Trailer & C.V. & Ped. & Motor. & Bicyc. & T.C. & Barrier \\ \midrule
\checkmark & & & 68.8 & 63.1 & 80.5 & 54.3  & 72.1 & 34.6 & \cellcolor{gray!15}29.0 & \cellcolor{gray!15}85.4 & \cellcolor{gray!15}72.3 & \cellcolor{gray!15}68.8 & \cellcolor{gray!15}78.0 & \cellcolor{gray!15}64.0 \\
& \checkmark & & 68.7 & 62.5 & \cellcolor{gray!15}83.9 & \cellcolor{gray!15}56.4 & \cellcolor{gray!15}73.4 & \cellcolor{gray!15}41.4 & 27.4 & 84.3 & 69.5 & 55.6 & 72.4 & 60.7 \\
\checkmark & \checkmark & & 71.5 & 68.6 & 85.2 & 60.2 & 75.0 & 41.5 & 32.9 & 87.4 & 77.2 & 72.0 & 80.1 & 74.1 \\
\checkmark & \checkmark & \checkmark & 72.7 & 70.4 & 86.1 & 62.5 & 76.8 & 44.8 & 34.4 & 88.6 & 78.7 & 73.7 & 82.6 & 75.5 \\ \bottomrule
\end{tabular}%
}
\end{center}
\caption{Ablation study of the Instance Generation module and Shape Alignment module. 
C.V.: Construction Vehicle. Ped.: Pedestrian. Motor.: Motorcycle. Bicyc.: Bicycle. T.C: Traffic Cone. 
\colorbox{gray!15}{Gray Cell}: By comparing the results generated by the LiDAR-instance-only model and camera-instance-only model, we highlight the results in which one performed better than the other.}
\label{tab:abl_lidar_cam_query}
\end{table*}

To reveal the pros and cons of the two kinds of instances, we remove each of them from \name for ablation.
To achieve this, we only choose instances from a single modality as the input of the Bi-modal Instance-based Prediction module.
There are several intriguing findings in Table~\ref{tab:abl_lidar_cam_query}:
\begin{itemize}
    \item 
    The model utilizing solely camera instances demonstrates superior performance in detecting objects of relatively small dimensions compared to the model that relies exclusively on LiDAR instances. Notably, the camera-instance-only model exhibits considerable superiority in identifying objects in categories such as Motorcycle, Bicycle, and Traffic Cone. Conversely, for categories encompassing larger-sized objects, like Car, Bus, and Trailer, the model that only uses  LiDAR instances outperforms the camera-instance-only model.
    
    \item 
    Combining these two kinds of instances, the performance is improved in all categories.
    It is noteworthy that the performance is significantly improved in some classes where the LiDAR-instance-only model and camera-instance-only model have similar performance, such as Bus and Pedestrian.
    This proves the complementarity of these bi-modal instances.
\end{itemize}

\subsubsection{Instance Shape Alignment}
To verify the effectiveness of our Shape Alignment module, we design a model without shape alignment, which we refer to as \namenospace-M (\namenospace-Misaligned).
Specifically, we directly use the reference boxes in Figure~\ref{fig:framework} as the final results. 
The last two rows of Table~\ref{tab:abl_lidar_cam_query} show the comparison between \namenospace-M and \name, where refining instances give us a 1.8 mAP boost.
Without the shape alignment, the performance degrades in all categories consistently, which indicates that shape alignment has an essential impact on the performance.

\subsubsection{Instance Feature Interaction}
In this section, we examine the role of the Instance Feature Interaction module in \S\ref{sec:inst_feat_interaction}.
Specifically, since the inputs and outputs of this module are the same (i.e., instance features), we can conduct the ablation study by simply removing this module.
As Table~\ref{table:abl_interaction_module} shows, the instance feature interaction module yields a 0.2 mAP and 0.6 NDS improvement.
In detail, this module notably decreases orientation error (from 0.345 to 0.317) and velocity error (from 0.244 to 0.218). 
\begin{table}[!h]
\begin{center}
\resizebox{\columnwidth}{!}{
\begin{tabular}{c|cc|ccccc}
\toprule
Interaction & mAP~$\uparrow$ & NDS~$\uparrow$ & mATE~$\downarrow$ & mASE~$\downarrow$ & mAOE~$\downarrow$ & mAVE~$\downarrow$ & mAAE~$\downarrow$ \\
\midrule
$\times$       & 70.2 & 72.1 & 0.279 & 0.247 & 0.345 & 0.244 & 0.187 \\
$\checkmark$   & 70.4 & 72.7 & 0.277 & 0.247 & 0.317 & 0.218 & 0.186 \\
\bottomrule
\end{tabular}
}
\end{center}
\caption{Ablation study of instance feature interaction module.}
\label{table:abl_interaction_module}
\end{table}

\subsubsection{Two-stage Assignment}
To verify the effectiveness of the two-stage assignment, we make comparisons between our strategy with different assignment strategies: Hungarian assignment~\cite{hungarian_assign}, Point-in-box assignment~\cite{fsd}, LiDAR-based assignment, LiDAR-based assignment with the BEV-distance-based assignment (introduced later) and our proposed two-stage assignment.
Specifically, for the Hungarian assignment, we adopt the implementation from DETR3D~\cite{detr3d}.
The point-in-box strategy corresponds to the assignment used in FSD~\cite{fsd}.
The LiDAR-based assignment is the upgraded version of the point-in-box assignment with the help of Eq.~\ref{eq:weighted_center}.
For the BEV-distance-based assignment, different from the camera-based assignment that uses 2D similarity, this assignment utilizes 3D similarity, defined as the reciprocal of the distance between the center of an instance and the center of a GT box in BEV space.
Then, we replace our original two-stage assignment with the alternative approach. 
These replacements are applied in both the generation of reference boxes in \S\ref{sec:inst_shape_refine} and final boxes in \S\ref{sec:inst_pred}.
The results are shown in Table~\ref{tab:abl_query_assignment}.
We reach these conclusions.
\begin{itemize}
    \item 
    The model adopting the Hungarian Assignment fails to converge.
    The reason is as follows:
    FSF generates camera instances and LiDAR instances concurrently.
    So an object may concurrently correspond to a LiDAR instance and a camera instance.
    We denote these two instances as a LiDAR-camera instance pair.
    The instances in a pair tightly overlap with each other.
    Hungarian Assignment chooses one instance in a pair as positive and the other one is regarded as negative.
    This confuses the classification head since these two instances actually represent a single object and make the model fail to converge.

    \item 
    The model using point-in-box assignment converges but shows an inferior performance to the LiDAR-based assignment due to the disturbance of irrelevant background points.

    \item 
    Performing a BEV-distance-based assignment in the second stage helps a bit. 
    However, it shows an inferior performance to our proposed two-stage assignment.
    This is because it regards a large number of camera instances as negative as \S\ref{sec:query_assign} discussed.
   
\end{itemize}
\begin{table}[]
\centering
\resizebox{0.95\columnwidth}{!}{%
\begin{tabular}{cc|c|c}
\toprule
\multicolumn{2}{c|}{Assignment}                     & \multirow{2}{*}{mAP}           & \multirow{2}{*}{NDS}           \\
First Stage             & Second Stage              &                                &                                \\ \midrule
Hungarian               & -                         & Not converged                  & Not converged                  \\
Point-in-box & -                         & 68.7                           & 71.9                           \\
LiDAR-based   & -                         & 69.0                           & 72.1                           \\
LiDAR-based   & BEV-dist.-based  & 69.2                           & 72.2                           \\
LiDAR-based   & Camera-based    & \textbf{70.4} & \textbf{72.7} \\ \bottomrule
\end{tabular}
}
\vspace{2mm}
\caption{Ablation study of different assignment methods. This experiment reveals the superiority of our proposed assignment.
BEV-dist.-based: BEV-distance-based.}
\label{tab:abl_query_assignment}
\end{table}

\subsubsection{Robustness against 2D Predictions}
It would be a major concern the model is robust with respect to the 2D part.
To answer this question, we change the default HTC to a basic Mask RCNN whose 2D Mask AP on nuImages is only 38.4.
As Table~\ref{tab:abl_2d_segmentor} shows, although the 2D performance has a significant drop after degrading the 2D part, the 3D performance of \name only has 1 mAP drop.
This is reasonable.
The mission of the 2D part is to detect the objects that are ignored by the LiDAR-only baseline. 
While the 2D precision significantly decreases, the 2D model maintains a high recall rate, effectively addressing the primary issues associated with a LiDAR-only baseline.
Besides, we also conduct an experiment that degrades 2D masks to 2D boxes for producing frustum. 
The result shows that downgrading the masks to boxes has minimal impact on 3D detection.
Representing an object as a mask or box only affects the shape of the camera instance.
However, our shape alignment module effectively solves the problem associated with the shapes.
\begin{table}[h]
\begin{center}
\resizebox{\columnwidth}{!}{%
\begin{tabular}{@{}l|cc|c|cc@{}}
\toprule
\multirow{2}{*}{Method}                               & \multicolumn{2}{c|}{2D Performance}            & \multicolumn{1}{l|}{\multirow{2}{*}{\makecell{Mask/Box}}} & \multicolumn{2}{c}{3D Performance} \\
                                                      & Mask$^\ast$              & Box$^\ast$              & \multicolumn{1}{l|}{}                             & mAP              & NDS             \\ \midrule
Mask RCNN~\cite{maskrcnn}       & 38.4                  & 47.8                  & Mask                                              & 69.4             & 72.3            \\ \midrule
\multirow{2}{*}{HTC~\cite{htc}} & \multirow{2}{*}{46.4} & \multirow{2}{*}{57.3} & Box                                               & 69.8             & 72.5            \\
                                                      &                       &                       & Mask                                              & 70.4             & 72.7            \\ \bottomrule
\end{tabular}
}
\end{center}
\caption{Ablation study of different 2D models. 
2D performance is evaluated on nuImage while 3D performance is evaluated on nuScenes.
Mask$^\ast$: Mask mAP.
Box$^\ast$: Box mAP.
The result reveals that our method is robust against the 2D models.}
\label{tab:abl_2d_segmentor}
\end{table}

\subsubsection{The Effect of Image Features}
\label{sec:use_img_feat}
To investigate the effectiveness of image features, we conducted an experiment in the following manner.
First, we pre-save the mask result generated by HTC.
Then each point is painted with the image feature derived from the image backbone.
Then, the image backbone is supervised by the losses of \name.
As illustrated in Table~\ref{tab:abl_img_feat}, employing image features yields a minor improvement of 0.3 mAP in our methodology. 
Training image backbones consumes a significant amount of time and GPU memory.
To ensure the efficiency of \name, we do not use the image feature in our method.
\begin{table}[H]
\begin{center}
\resizebox{\columnwidth}{!}{%
\begin{tabular}{c|cc|ccccc}
\toprule
Image features & mAP~$\uparrow$  & NDS~$\uparrow$  & mATE~$\downarrow$  & mASE~$\downarrow$  & mAOE~$\downarrow$  & mAVE~$\downarrow$  & mAAE~$\downarrow$  \\ \midrule
\textbf{$\times$} & 70.4 & 72.7 & 0.277 & 0.247 & 0.317 & 0.218 & 0.186 \\
\checkmark        & 70.7 & 72.9 & 0.274 & 0.247 & 0.307 & 0.223 & 0.187 \\ \bottomrule
\end{tabular}
}
\end{center}
\caption{Ablation study of using the image feature. 
Utilizing image features does not result in a significant improvement.
As a result, we make it an optional choice.}
\label{tab:abl_img_feat}
\end{table}

\subsection{Comparison with Alternatives to FSF}\label{sec:abl_fsd_fusion}
In this section, we reproduce several alternatives that can be adapted for fully sparse architecture and conduct a comprehensive comparison between them and FSF.

\subsubsection{Score Painting}
The most straightforward approach is using PointPainting~\cite{pointpainting}, where the 2D semantic scores are painted onto points. 
The results in Table~\ref{tab:abl_fsd_fusion} show simple point painting could obtain considerable improvement but is largely inferior to FSF.

\subsubsection{Feature Painting}
Feature painting~\cite{pointaugmenting, deepfusion, autoalign} paints image features to the point cloud.
Features contain richer information than semantic scores.
We adapt the painting strategy in PointAugmenting~\cite{pointaugmenting} on FSD.
Compared with score painting, feature painting has an improvement of 0.4 mAP.

\subsubsection{Virtual Point}
MVP~\cite{mvp} utilizes virtual points to augment the original point cloud.
Specifically, MVP utilizes 2D instance segmentation to generate 2D masks. 
Next, MVP uniformly samples 2D points in a 2D mask.
Then, it projects the LiDAR points onto the 2D image plane.
Each sampled 2D point finds the nearest projected LiDAR point and uses the depth of this LiDAR point as its own depth.
Finally, given the depth, the sampled 2D point is unprojected to 3D space and augments the original point cloud.
Then the augmented point clouds are sent into a LiDAR-based detector.
For a fair comparison, we use the official code to generate virtual points and then feed the augmented point cloud into FSD.
Table~\ref{tab:abl_fsd_fusion} demonstrates that adopting MVP achieves better performance than the painting method above, but is still worse than ours.

\subsubsection{Discussion: why is FSF superior to the alternatives?}
Although all three alternatives achieve improvement on the LiDAR-only baseline, they are significantly worse than FSF.
We owe the superior performance of FSF to two aspects:
\begin{itemize}
    \item 
    % The painting approaches utilize semantic scores or image features.
    Semantic scores and image features are rich in semantic-level information but lack instance-level information. 
    However, instance-level information is vital to detection since it greatly helps recognize objects within the same category.
    \item The virtual point approach employs 2D instance segmentation.
    However, it only augments the point cloud. 
    The objects with few points are still likely to be missed out.
    Nevertheless, FSF ensures that a 2D mask will produce a camera instance even when an object contains only one point.
    This is greatly helpful in detecting objects with a minimal number of points.
\end{itemize}

\begin{table}[h]
\begin{center}
\resizebox{0.8\columnwidth}{!}{%
\begin{tabular}{l|c|cc}
\toprule
Method                 & Modality & mAP  & NDS   \\
\midrule
FSD~\cite{fsd}                    & L        & 62.5 & 68.7 \\ 
% \midrule
FSD + Score Painting~\cite{pointpainting}   & L+C      & 66.9 & 70.9 \\
FSD + Feature Painting~\cite{pointaugmenting} & L+C      & 67.3 & 71.2 \\
FSD + Virtual Point~\cite{mvp}              & L+C      & 67.6 & 71.5 \\
FSF            & L+C      & 70.4 & 72.7 \\ \bottomrule
\end{tabular}%
}
\end{center}
\caption{Different fusion strategies for sparse detection architecture. 
Score Painting~\cite{pointpainting}: painting semantic scores. 
Feature Painting~\cite{pointaugmenting}: painting the image features. 
Virtual point~\cite{mvp}: employing virtual points to augment point cloud. 
}
\label{tab:abl_fsd_fusion}
\end{table}

\subsection{Long Range Detection}
\name achieves state-of-the-art performance on the Argoverse 2 dataset, which has a very large perception area ($400m \times 400m$).
We conduct the following experiments to showcase the superiority of FSF in long-range detection. 

\subsubsection{Range-conditioned Performance}
% Table~\ref{tab:abl_argo_dist} demonstrates the performance conditioned on different perception ranges.
The image is dense and of high resolution, making it helpful for detecting distant objects.
We conduct the range-conditioned performance to validate the effectiveness of our method.
Table~\ref{tab:abl_argo_dist} demonstrates the performance conditioned on different perception ranges.
As Table~\ref{tab:abl_argo_dist} shows, FSF greatly enhances the capabilities of FSD in long-range scenarios.
Particularly, detecting small objects at a long distance is especially challenging, which is a good testbed to showcase the effectiveness of FSF.
Therefore, we select three small object categories that are frequently encountered: Motorcyclist, Bicyclist as representatives.
In the range spanning from 50m to 100m, we observe a remarkable improvement in mAP for these small objects.
\begin{table}[h]
\begin{center}
\resizebox{\columnwidth}{!}{%
\begin{tabular}{l|c|cccc|cccc}
\toprule
\multirow{2}{*}{} & \multirow{2}{*}{\makecell{Overall \\ mAP}} & \multicolumn{4}{c|}{0m-50m}                                       & \multicolumn{4}{c}{50m-100m}                                      \\  
                  &                              & \multicolumn{1}{c}{Avg.}  & Motor. & Bicyc. & C.B. & \multicolumn{1}{c}{Avg.}  & Motor. & Bicyc. & C.B. \\ \midrule
FSD        & 28.1                         & \multicolumn{1}{c}{41.6} & 57.3        & 57.4      & 66.1      & \multicolumn{1}{c}{10.9} & 8.8         & 17.0      & 13.8      \\
FSF               & 33.2                         & \multicolumn{1}{c}{45.3} & 57.9        & 65.1      & 73.7      & \multicolumn{1}{c}{17.2} & 36.4        & 31.1      & 25.2      \\ \bottomrule
\end{tabular}%
}
\end{center}
\caption{Performance of different ranges on Argoverse 2. 
FSF significantly improves the performance on faraway objects relative to the FSD~\cite{fsd}.
Avg.: the average mAP of all classes.
Motor.: Motorcyclist. 
Bicyc.: Bicyclist.
C.B.: Construction Barrel.
We select three representative categories of small objects for demonstration.
}
\label{tab:abl_argo_dist}
\end{table}

\subsubsection{Latency and Memory Footprint Analysis}
Here we emphasize the superiority of \name in long-range detection by evaluating the inference latency and memory footprint.
The latency is tested on RTX 3090 with batch size 1. Our code is based on MMDetection3D~\cite{mmdet3d} and the IO time is excluded.
For comparison, we use the officially released codebases of CenterPoint and TransFusion to obtain their results.

First, we introduce the specific settings in Table~\ref{tab:argo_latency_mem}: 
\begin{itemize}
\item \textbf{Image Process.}
In TransFusion, the image component is constructed on a ResNet50 backbone and operates on images of size $800 \times 320$. FSF adopts the same backbone and image size as TransFusion but extends the image backbone with a 2D head for generating camera instances.

\item \textbf{LiDAR Process.}
CenterPoint, TransFusion-L, and TransFusion employ a VoxelNet-like~\cite{voxelnet} LiDAR backbone followed by 2D convolutions to generate BEV feature maps.
FSD and FSF use the SparseUNet~\cite{parta2} backbone to extract voxel features.
Then, unlike BEV-based methods, FSD and FSF construct the point features from voxel features.
These point features are used  to produce LiDAR instances.

\item \textbf{Prediction Process.}
We define the prediction process as the process of generating 3D bounding boxes from a BEV feature map or 3D instances.
The prediction process of each method is as follows: 

(a) CenterPoint uses an MLP head to produce 3D boxes from the features of BEV grids.

(b) TransFusion operates in three steps:
    (i) It first utilizes an MLP head to classify the features of BEV grids and then selects top-k grids to form k queries.
    (ii) TransFusion then utilizes a transformer decoder with global cross-attention to update the feature of queries.
    The global attention is performed between queries and the feature maps from both image and LiDAR modalities. 
    (iii) It finally outputs 3D bounding boxes based on the features of queries using an MLP head.
    
(c) TransFusion-L is identical to TransFusion but omits the step of performing cross-attention between queries and image feature maps.

(d) FSD and FSF first extract instance features from instances using the SIR module.
Then they employ these instance features to generate reference boxes for shape alignment.
Finally, FSD and FSF generate 3D bounding boxes from the shape-aligned instances.
\end{itemize}

As Table~\ref{tab:argo_latency_mem} shows, our FSF method shows a remarkable advantage in both latency and memory footprint over other state-of-the-art multi-modal methods. 
It is mainly attributed to two reasons:
\begin{itemize}
    \item FSF does not construct a BEV feature map. 
    Considering a detection area of $400m \times 400m$, creating a BEV feature map over such an extended range would require 16 times more memory compared to a $100m \times 100m$ area. Moreover, conducting 2D convolution operations on this BEV feature map would be computationally intensive and time-consuming.
    \item FSF has a much faster prediction process than TransFusion.
    Transfusion needs to perform global attention with the BEV feature map and image feature maps. 
    Meanwhile, FSF directly generates bounding boxes from 3D instances, bypassing the need to process features of the background.
\end{itemize}

Despite the need to process image modality, our approach remains faster and requires less memory than well-known LiDAR-only methods such as CenterPoint~\cite{centerpoint}.
\begin{table}[ht]
\begin{center}
\resizebox{\columnwidth}{!}{%
\begin{tabular}{l|c|cccc|c}
\toprule
\multirow{2}{*}{Method} & \multirow{2}{*}{Modality} & \multicolumn{4}{c|}{Latency (ms)} & \multirow{2}{*}{Memory (GB)} \\ 
                        &                           & Image & LiDAR & Prediction & Overall &                            \\ \midrule
TransFusion-L           & L                         & -     & 198    & 122       & 320     & 14.5                       \\
CenterPoint             & L                         & -     & 190    & 42        & 232     & 8.1                        \\
FSD                     & L                         & -     & 56     & 41        & \textbf{97}  & \textbf{3.8}                \\ \midrule
TransFusion             & C+L                       & 30    & 198    & 156       & 384     & 17.3                       \\
FSF (ours)              & C+L                       & 41    & 56     & 44        & \textbf{141} & \textbf{6.9}                \\ \bottomrule
\end{tabular}%
}
\end{center}
\caption{Inference latency and memory footprint comparison on the Argoverse 2 dataset, covering a $400m \times 400m$ area.}
\label{tab:argo_latency_mem}
\end{table}

\begin{figure*}[!ht]
\centering
\includegraphics[width=1.0\textwidth]{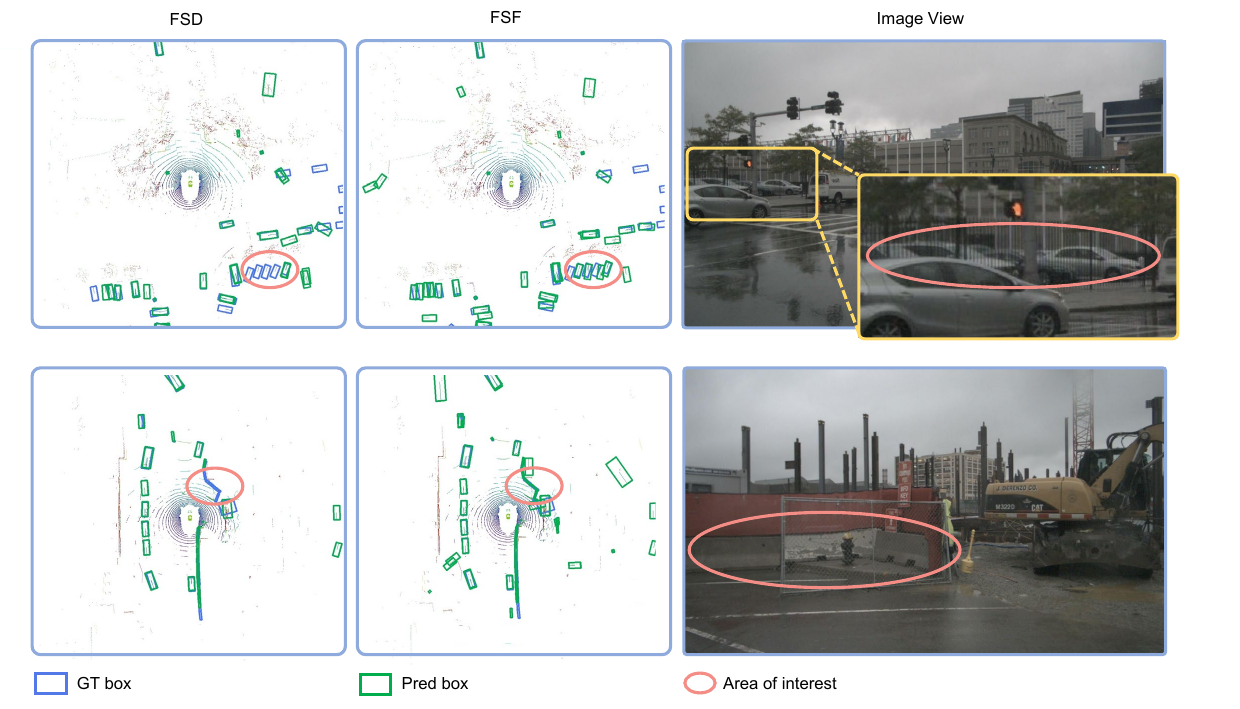}
\caption{
Qualitative comparison between FSD~\cite{fsd} and FSF.
In these scenes, some objects are partially occluded by wire netting, leading to very few LiDAR points.
We mark the areas of interest with \red{red} circles. 
With the help of image information, FSF detects more objects than FSD. 
}
\label{fig:appendix_vis}
\end{figure*}

\subsubsection{Latency Breakdown}
We have provided a more detailed runtime analysis of each component in Table~\ref{tab:detailed_latency}. 
Despite a somewhat long pipeline of FSF, each component is designed to be efficient, ensuring that the overall processing speed remains rapid.
\begin{table}[h]
\begin{center}
\begin{tabular}{l|c}
\toprule
Module Name & Latency (ms) \\
\midrule
\textbf{Camera Instance Generation} & \textbf{41.1} \\
\quad 2D Instance Segmentation & 39.8 \\
\quad Point Cropping & 1.3 \\
\midrule
\textbf{LiDAR Instance Generation} & \textbf{56.2} \\
\quad LiDAR Backbone & 42.3 \\
\quad Point Classification + Center Voting + Connecting & 7.8 \\
\quad Grouping by CCL & 6.1 \\
\midrule
\textbf{Bi-modal Instance-based Prediction} & \textbf{43.9} \\
\quad Instance Feature Extraction 1 & 9.2 \\
\quad Self-Attn 1 & 5.8 \\
\quad Head 1 & 1.8 \\
\quad Point Cropping & 2.2 \\
\quad Feature Extraction 2 & 9.2 \\
\quad Self-Attn 2 & 5.8 \\
\quad Head 2 & 1.8 \\
\quad NMS & 8.1 \\
\bottomrule
\end{tabular}
\end{center}
\caption{Breakdown of FSF latency on the Argoverse2 dataset using the RTX 3090 GPU.
The order of the modules in the table corresponds to their arrangement in Fig.~\ref{fig:framework}.}
\label{tab:detailed_latency}
\end{table}

\subsection{Evaluating FSF and FSD in Hard Scenarios}
In Fig.~\ref{fig:inst_seg}, we hypothesize that multi-modal FSF can enhance the performance of single-modal FSD in two scenarios. 
The first is when points are very few. 
The second is when numerous objects are closely packed together.
To validate our claims, we conducted the following experiments. 
\subsubsection{Superior Performance for Sparsely-Pointed Objects}
We first demonstrate the performance improvement of FSF over FSD under the setting of different numbers of points within objects. 
We categorize objects based on the number of points they contain, splitting them into four groups: $(0, 10]$, $(10, 30]$, $(30, 100]$, and $(100, \infty)$. 
Since mAP is calculated at the scene level and each scene contains objects with varying numbers of points, directly evaluating mAP for objects with a specific number of points becomes problematic.
Therefore, we adopt \emph{mean average recall} as the metric of Table~\ref{tab:pts_num_recall}.
Across all categories, FSF consistently outperforms FSD. Notably, in the $(0, 10]$ group, the improvement is most pronounced, showing an overall increase of 4.3 recall. For small objects like bicycles and traffic cones, which contain very few points, pure 3D segmentation struggles with recognition. However, with the assistance of 2D segmentation, identification is simplified (10 recall boost).
\begin{table}[h]

\begin{center}

\resizebox{0.95\columnwidth}{!}{

\begin{tabular}{l|c|c|c|ccc}

\toprule

Method & \#Points & \#GT boxes & mAR & Motor. & Bicyc. & T.C. \\ \midrule

FSD & \multirow{2}{*}{0-10} & \multirow{2}{*}{59457} & 52.3 & 61.3 & 69.4 & 82.1  \\

FSF & & & 56.6~\up{4.3} & 67.7 & 79.6\up{10.2} & 91.7\up{9.6} \\\midrule

FSD & \multirow{2}{*}{10-30} & \multirow{2}{*}{25735} & 69.6 & 93.1 & 88.2 & 96.1 \\

FSF & & & 71.9~\up{2.3} & 96.9 & 92.8 & 98.9 \\\midrule

FSD & \multirow{2}{*}{30-100} & \multirow{2}{*}{18865} & 78.1 & 97.1 & 94.2 & 98.1 \\

FSF & & & 80.5~\up{2.4} & 98.9 & 98.8 & 98.9 \\\midrule

FSD & \multirow{2}{*}{100-inf} & \multirow{2}{*}{17804} & 85.2 & 98.1 & 94.2 & 97.1 \\

FSF & & & 87.3~\up{1.9} & 100.0 & 100.0 & 98.9 \\

\bottomrule

\end{tabular}%

}

\end{center}

\caption{Comparison of \emph{recall} values under the different numbers of  points within objects. 
The recall threshold is $0.5m$. 
Motor.: Motorcyclist. 
Bicyc.: Bicyclist.
T.C.: Traffic Cone.
We select three representative categories of small objects for demonstration.}

\label{tab:pts_num_recall}

\end{table}

\subsubsection{Superior Performance in Crowded Scenarios}
This experimental section evaluates the FSD and FSF performance in crowded scenarios. 
In crowded scenarios, 3D segmentation is prone to group multiple objects as a single instance, whereas 2D segmentation is skilled at differentiating between these objects.

To better understand the challenges posed by crowded scenarios, we define two terminologies: \emph{crowded box} and \emph{crowded scene}.
Considering all the boxes in the BEV perspective, a GT BEV box \(B^*_i\) is designated as a crowded box if its 2D center lies within a distance threshold \(\varepsilon_i\) from the 2D center of another GT box \(B^*_j\). 
The threshold \(\varepsilon_i\) is defined as the diagonal length of \(B^*_i\). 
Thus, if the center-to-center distance between \(B^*_i\) and \(B^*_j\) falls below \(\varepsilon_i\), \(B^*_i\) qualifies as a \emph{crowded box}.
A \emph{crowded scene} is identified when over \(50\%\) of the GT boxes in this scene are \emph{crowded boxes}.

In Table~\ref{tab:crowd_scenes}, it is evident that the FSD method experiences a notable performance drop in crowded scenes compared to overall scenarios, with a mAP reduction of 6.8. 
This decline is primarily attributed to the limitations of 3D instance segmentation in accurately distinguishing objects that are closely packed together. 
In contrast, our proposed FSF method, which employs 2D instance segmentation, has enhanced capability to identify these objects separately. 
Specifically, FSF achieves a lower mAP drop of 3.6.
FSF has an 11.1 mAP improvement over FSD in crowded scenes, a more significant gain compared to the 7.9 mAP increase observed in overall scenarios.

\begin{table}[h]
\begin{center}
\resizebox{0.95\columnwidth}{!}{
\begin{tabular}{l|c|c|cccc}
\toprule
Method & Scene Type & NDS & mAP & Motor. & Bicyc. & T.C. \\ \midrule
FSD & crowded & 65.2 & 55.7 & 66.0 & 44.9 & 55.0 \\
FSF & crowded & 70.4 & 66.8~\up{11.1} & 77.6 & 69.5 & 72.6 \\ \midrule
FSD & all& 68.7 & 62.5 & 69.5 & 55.6 & 72.5 \\
FSF & all& 72.7 & 70.4~\up{7.9} & 78.7 & 73.7 & 82.6 \\ \bottomrule
\end{tabular}
}
\end{center}
\caption{Comparison for FSD and FSF across crowded scenes and all scenes. 
Motor.: Motorcyclist. 
Bicyc.: Bicyclist.
T.C.: Traffic Cone.
We select three representative categories of small objects for demonstration.
}
\label{tab:crowd_scenes}
\end{table}

\subsection{Qualitative Results}
To gain insight into the inner workings of \namenospace, we present the qualitative results. 
An interesting case is demonstrated in Figure~\ref{fig:appendix_vis}. 
Here, FSD~\cite{fsd} fails to detect objects that are partially occluded by wire netting. 
Due to the wire netting, a significant portion of the lasers emitted by LiDAR is intercepted, resulting in few point clouds on these objects. 
It is difficult for LiDAR-only methods to identify these objects. 
However, using images, the masks of these objects are easily identified, allowing \name to detect them successfully.

\section{Conclusion}
We present the Fully Sparse Fusion (FSF) framework, the first multimodal architecture designed for full sparsity. 
FSF is an instance-level fusion method that contains no dense BEV feature maps. 
We introduce a Bi-modal Instance Generation module that performs joint 2D and 3D instance segmentation to produce bi-modal instances. 
Subsequently, these instances become shape-aligned and are used for bounding box prediction through our Bi-modal Instance-based Prediction Module. 
Moreover, an effective Bi-modal Assignment Strategy is proposed to properly assign these bi-modal instances.
Our FSF framework sets new state-of-the-art performance on three large-scale datasets: nuScenes, WOD, and Argoverse 2, covering both short-range and long-range scenarios. 
Notably, FSF exhibits exceptional capability and efficiency in long-range detection, significantly reducing latency and memory overhead.

{\appendices
% \begin{figure}[h]
% \centering
% \includegraphics[width=1.0\columnwidth]{image/Recall.pdf}
% \caption{Recall comparison in metric 0.5m. 
% C.V.: Construction Vehicle. Ped.: Pedestrian. Motor.: Motorcycle. Bicyc.: Bicycle. T.C: Traffic Cone. }
% \label{fig:recall}
% \end{figure}

% \section{Different Lighting}
% \input{tables/abl_day_night}

% \subsection{Implementation Details}
% \input{tables/imple_detail}

}

{
\bibliographystyle{IEEEtran}
\bibliography{egbib}
}

% \newpage

\begin{IEEEbiography}[{\includegraphics[width=1in,height=1.25in,clip,keepaspectratio]{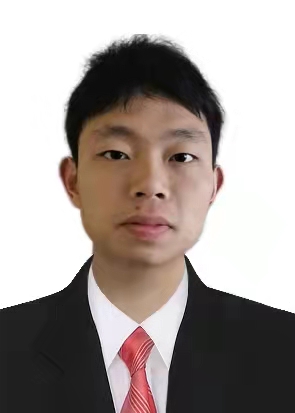}}]{Yingyan Li} received his BS degree from the University of Chinese Academy of Sciences, Beijing, in 2021.
He is currently a Ph.D. student in the National
Laboratory of Pattern Recognition (NLPR),
Institute of Automation, Chinese Academy of
Sciences (CASIA), Beijing, China, supervised
by Prof. Zhaoxiang Zhang. He is also currently
with the School of Future Technology, University
of Chinese Academy of Sciences (UCAS). 
His major
research interests include 3D perception and autonomous driving.
\end{IEEEbiography}

\begin{IEEEbiography}[{\includegraphics[width=1in,height=1.25in,clip,keepaspectratio]{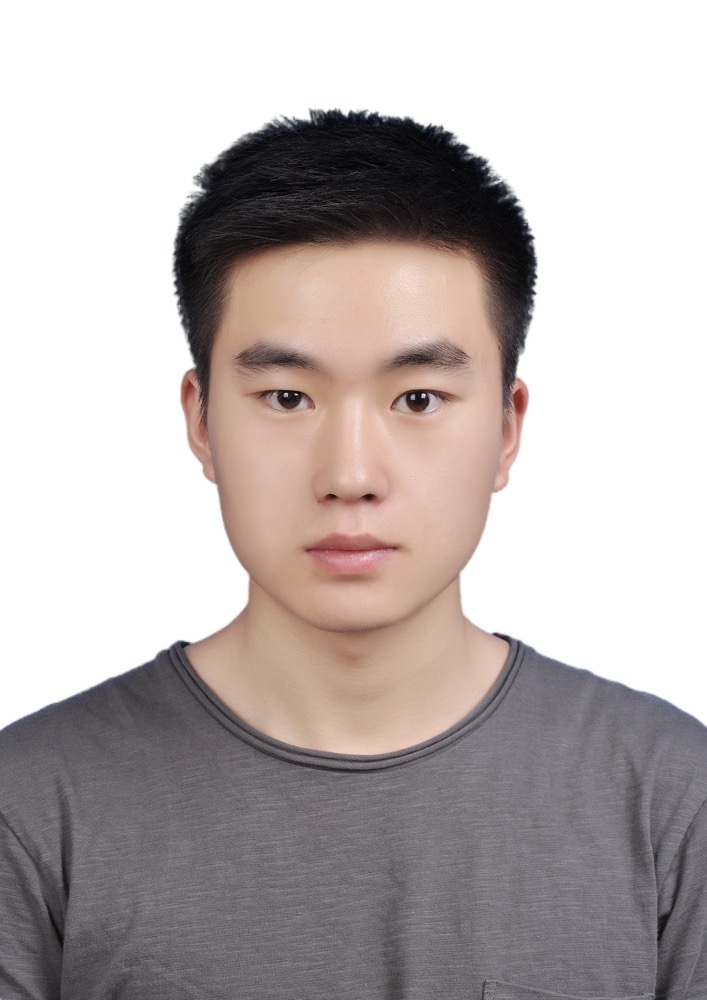}}]{Lue Fan}
is currently a Ph.D. student in the National
Laboratory of Pattern Recognition (NLPR),
Institute of Automation, Chinese Academy of
Sciences (CASIA), Beijing, China, supervised
by Prof. Zhaoxiang Zhang. He is also currently
with the School of Future Technology, University
of Chinese Academy of Sciences (UCAS). He
got his BS degree from Xi’an Jiaotong University
in 2019, and majored in automation. His major
research interests lie in 3D/2D perception algorithms
and applications.
\end{IEEEbiography}

\begin{IEEEbiography}[{\includegraphics[width=1in,height=1.25in,clip,keepaspectratio]{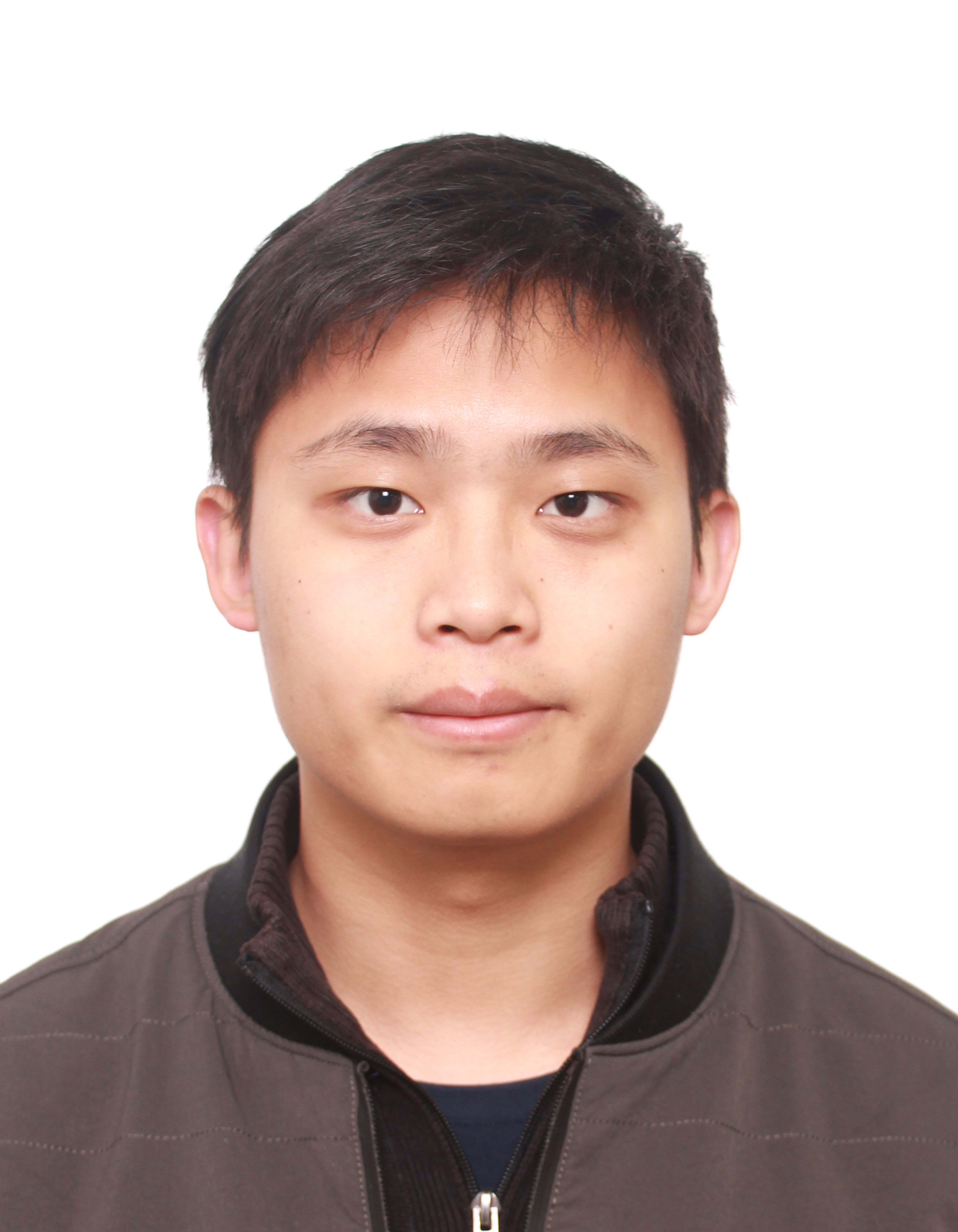}}]{Yang Liu}
received the BS degree in Yingcai Honor College from University of Electronic Science and Technology of China, in 2018. He is currently an PhD student at Institute of Automation, Chinese Academy of Sciences. His research interests include 3D computer vision and reconstruction. 
\end{IEEEbiography}

\begin{IEEEbiography}[{\includegraphics[width=1in,height=1.25in,clip,keepaspectratio]{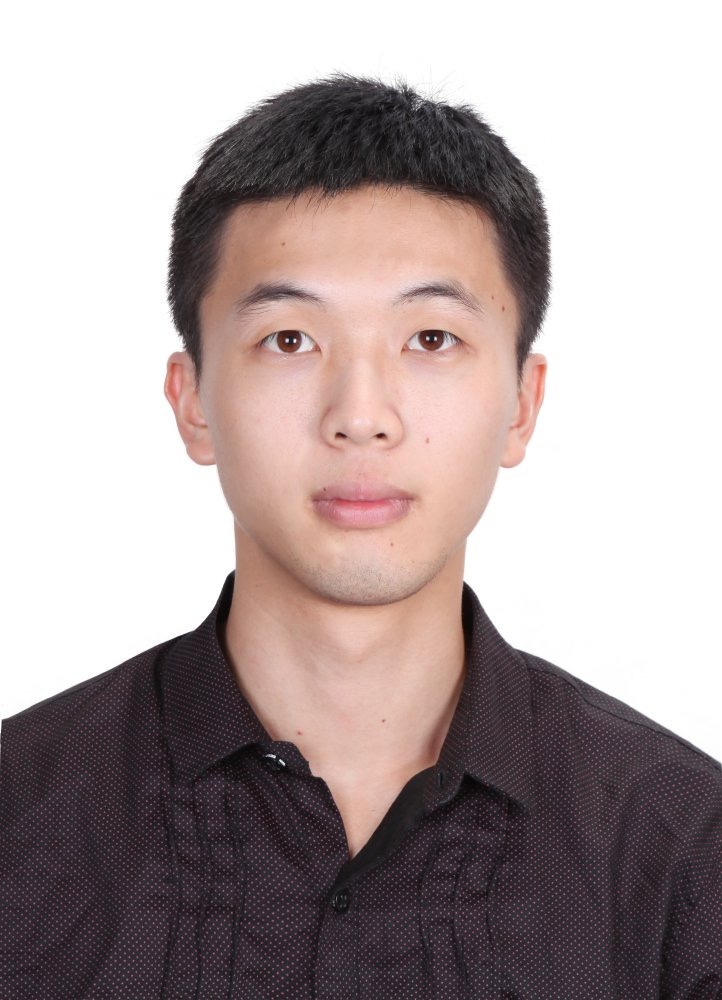}}]{Zehao Huang} received the BS degree in automatic control from Beihang University, Beijing,
China, in 2015. 
He is currently an algorithm engineer at TuSimple. His research interests include
computer vision and image processing.
\end{IEEEbiography}

\begin{IEEEbiography}[{\includegraphics[width=1in,height=1.25in,clip,keepaspectratio]{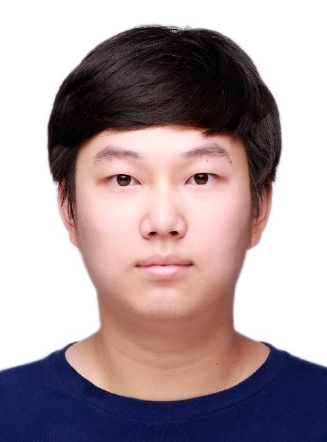}}]{Yuntao Chen} received his Bachelor degree in University of Science and Technology, Bejing, in 2016. He received the Ph.D. degree from the Center for Research on Intelligent Perception and Computing (CRIPAC), National Laboratory of Pattern Recognition (NLPR), Institute of Automation, Chinese Academy of Sciences (CASIA), Beijing, China, in 2021. His research interests include object recognition and 3D scene understanding.
\end{IEEEbiography}

\begin{IEEEbiography}[{\includegraphics[width=1in,height=1.25in,clip,keepaspectratio]{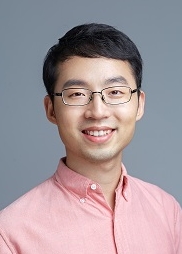}}]{Naiyan Wang}
is currently the chief scientist of
TuSimple. he leads the algorithm research group
in the Beijing branch. Before this, he got his Ph.D.
degree from the CSE department, HongKong
University of Science and Technology in 2015.
His supervisor is Prof. Dit-Yan Yeung. He got
his BS degree from Zhejiang University, in 2011
under the supervision of Prof. Zhihua Zhang.
His research interest focuses on applying the
statistical computational model to real problems
in computer vision and data mining. Currently, He
mainly works on the vision-based perception and localization part of
autonomous driving. Especially He integrates and improves the cuttingedge
technologies in academia, and makes them work properly in the
autonomous truck.
\end{IEEEbiography}

\begin{IEEEbiography}[{\includegraphics[width=1in,height=1.25in,clip,keepaspectratio]{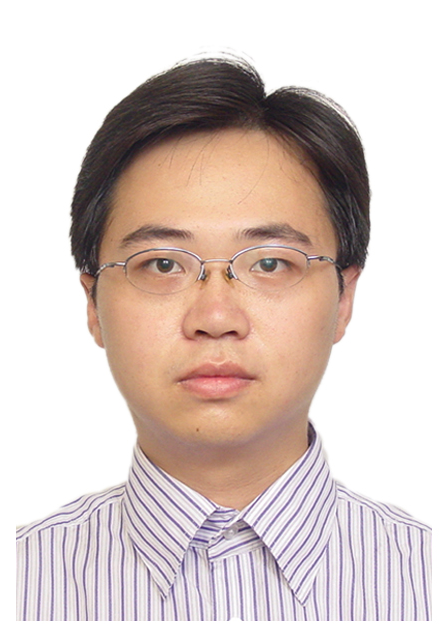}}]{Zhaoxiang Zhang}
received the bachelor’s degree
in circuits and systems from the University
of Science and Technology of China (USTC) in
2004 and the Ph.D. degree from the National Laboratory
of Pattern Recognition (NLPR), Institute
of Automation, Chinese Academy of Sciences
(CASIA), Beijing, China, in 2009. In October 2009,
he joined the School of Computer Science and
Engineering, Beihang University, and worked as
an Assistant Professor from 2009 to 2011, an
Associate Professor from 2012 to 2015, and the
Vice-Director of the Department of Computer Application Technology
from 2014 to 2015. In July 2015, he returned to the CASIA, to join as a
Professor, where he is currently a Professor with the Center for Research
on Intelligent Perception and Computing. He has published more than
200 papers in reputable conferences and journals. His major research
interests include pattern recognition, computer vision, machine learning,
and bio-inspired visual computing. He has won the best paper awards in
several conferences.
\end{IEEEbiography}

\end{document}